%% file: acl_latex.tex
\pdfoutput=1

\documentclass[11pt]{article}

\usepackage[final]{acl}

\usepackage{times}
\usepackage{latexsym}

\usepackage[T1]{fontenc}

\usepackage[utf8]{inputenc}

\usepackage{microtype}

\usepackage{inconsolata}

\usepackage{graphicx}
\usepackage{scalerel,xparse}
\usepackage{booktabs}
\usepackage{xcolor,colortbl}
\usepackage{multirow}
\usepackage{multicol}
\usepackage{subfigure}
\usepackage{floatrow}
\usepackage{pmboxdraw}
\usepackage{fontawesome5}
\usepackage{fancyvrb}
\usepackage[many]{tcolorbox}
\usepackage{tikz}
\usepackage{scalerel}
\usepackage{amssymb}
\usepackage{array}
\usepackage{cuted}
\usepackage[frozencache,cachedir=.]{minted}
\setminted[python]{breaklines}
\usepackage{pifont}
\usepackage{arydshln}
\usepackage{listings}
\tcbuselibrary{listingsutf8}
\usepackage[textsize=tiny]{todonotes}

\newcommand*\colourcheck[1]{%
  \expandafter\newcommand\csname #1check\endcsname{\textcolor{#1}{\ding{52}}}%
}
\colourcheck{green}
\colourcheck{blue}

\usepackage{enumitem}
\NewDocumentCommand\prism{}{
\includegraphics[scale=0.052]{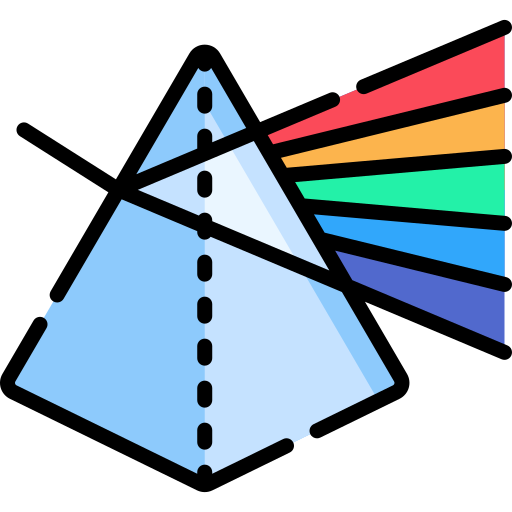}}

\NewDocumentCommand\prismsmall{}{
\includegraphics[scale=0.03]{Figures/prism_icon.png}}

\definecolor{lightblue}{rgb}{0.68, 0.85, 0.9}
\definecolor{bblue}{HTML}{5989cf}
\definecolor{LightGray}{gray}{0.9}

\newtcolorbox{doubbox}{
    enhanced, 
    boxrule = 0pt, 
    rounded corners,
    colback=white,
    borderline = {0.75pt}{0pt}{bblue}, 
    borderline = {0.75pt}{2pt}{lightblue} 
}
\newtcolorbox{singlebox}{
    enhanced, 
    boxrule = 0pt, 
    rounded corners,
    colback=white,
    borderline = {0.75pt}{0pt}{bblue}, 
}

\newtcolorbox{PromptBox}{
  breakable,
  enhanced,
  colback=red!5!white,
  colframe=red!75!black,
  fonttitle=\bfseries,
  title={Prompt for taxonomy annotation},
  width=\textwidth,
  listing only,
  listing engine=listings,
  listing options={
    basicstyle=\ttfamily\small,
    breaklines=true,
    breakatwhitespace=false,
    postbreak=\mbox{\textcolor{red}{$\hookrightarrow$}\space},
    columns=fullflexible,
    keepspaces=true,
    showstringspaces=false
  }
}

\title{\prism \ \textsc{PromptPrism}: A Linguistically-Inspired Taxonomy for Prompts}

\author{
\textbf{Sullam Jeoung},
\textbf{Yueyan Chen},  
\textbf{Shuai Wang},   
\textbf{Haibo Ding},   
\textbf{Yi Zhang},   
\textbf{Lin Lee Cheong}
\\  
Amazon AWS
\\
\small{
    \texttt{(sullamij, yyanc, wshui, hbding, yizhngn, lcheong)@amazon.com}
  }
}

\begin{document}
\maketitle
\begin{abstract}
Prompts are the interface for eliciting the capabilities of large language models (LLMs). Understanding their structure and components is critical for analyzing LLM behavior and optimizing performance. However, the field lacks a comprehensive framework for systematic prompt analysis and understanding. We introduce \prismsmall \textsc{PromptPrism}, a linguistically-inspired taxonomy that enables prompt analysis across three hierarchical levels: functional structure, semantic component, and syntactic pattern. By applying linguistic concepts to prompt analysis, \textsc{PromptPrism} bridges traditional language understanding and modern LLM research, offering insights that purely empirical approaches might miss. We show the practical utility of \textsc{PromptPrism} by applying it to three applications: (1) a taxonomy-guided prompt refinement approach that automatically improves prompt quality and enhances model performance across a range of tasks; (2) a multi-dimensional dataset profiling method that extracts and aggregates structural, semantic, and syntactic characteristics from prompt datasets, enabling comprehensive analysis of prompt distributions and patterns; (3) a controlled experimental framework for prompt sensitivity analysis by quantifying the impact of semantic reordering and delimiter modifications on LLM performance. Our experimental results validate the effectiveness of our taxonomy across these applications, demonstrating that \textsc{PromptPrism} provides a foundation for refining, profiling, and analyzing prompts. 

\end{abstract}

\input{0_introduction}

\input{1_related_works}
\input{2_prompt_prism}

\input{3_case_study}

\input{4_conclusion}

\bibliography{custom}

\appendix
\input{5_appendix}

\end{document}

%% file: 0_introduction.tex
\section{Introduction}

The systematic study of prompt design and optimization has emerged as a critical research area in the evolution of Large Language Models (LLMs) \cite{brown2020language,touvron2023llama, zhao2023survey}. While effective prompts can significantly enhance model performance, studies have revealed that LLMs demonstrate notable sensitivity to various prompt characteristics \cite{liu2024lost,sclarquantifying,li2024think}, including stylistic formatting, sequential ordering, task explanations, and meta-rules. This sensitivity, coupled with inconsistent documentation of prompt specifications across the literature, presents challenges for systematic performance evaluation and comparative analysis. 


Despite the fundamental importance of prompting in LLM interactions, there exists a notable gap in the development of frameworks that enable comprehensive analysis of prompts themselves. Previous approaches focus on analyzing generated responses along dimensions such as helpfulness and harmlessness \cite{askell2021general}, while systematic understanding of prompt characteristics remains understudied. While recent works have proposed standardized documentation protocols to enhance prompt transparency  \citep{bach2022promptsource,wang2022super}, these efforts often lack the depth and structure needed to capture the nuanced variations in prompt composition and their effects on model behavior.

Linguistic theory offers a valuable foundation for addressing this gap \cite{opitz2025natural}. Far from being merely a nostalgic endeavor, linguistic approaches provide established analytical frameworks for understanding how language functions as a structured communication system \cite{ravichander2022condaqa, ponti2020xcopa}. These principles remain relevant in the era of large language models precisely because LLMs, despite their neural architecture, fundamentally process and generate language. The historical application of linguistics in NLP has demonstrated that linguistic expertise is crucial for developing high-quality resources and evaluation methods \cite{opitz2025natural} -- from selecting diverse and representative language data \cite{ravichander2022condaqa} to ensuring proper annotation standards and designing effective evaluation metrics \cite{chi2024modeling, parrish2021bbq}. 



\begin{figure*}[h]
    \centering
    \includegraphics[width=\textwidth]{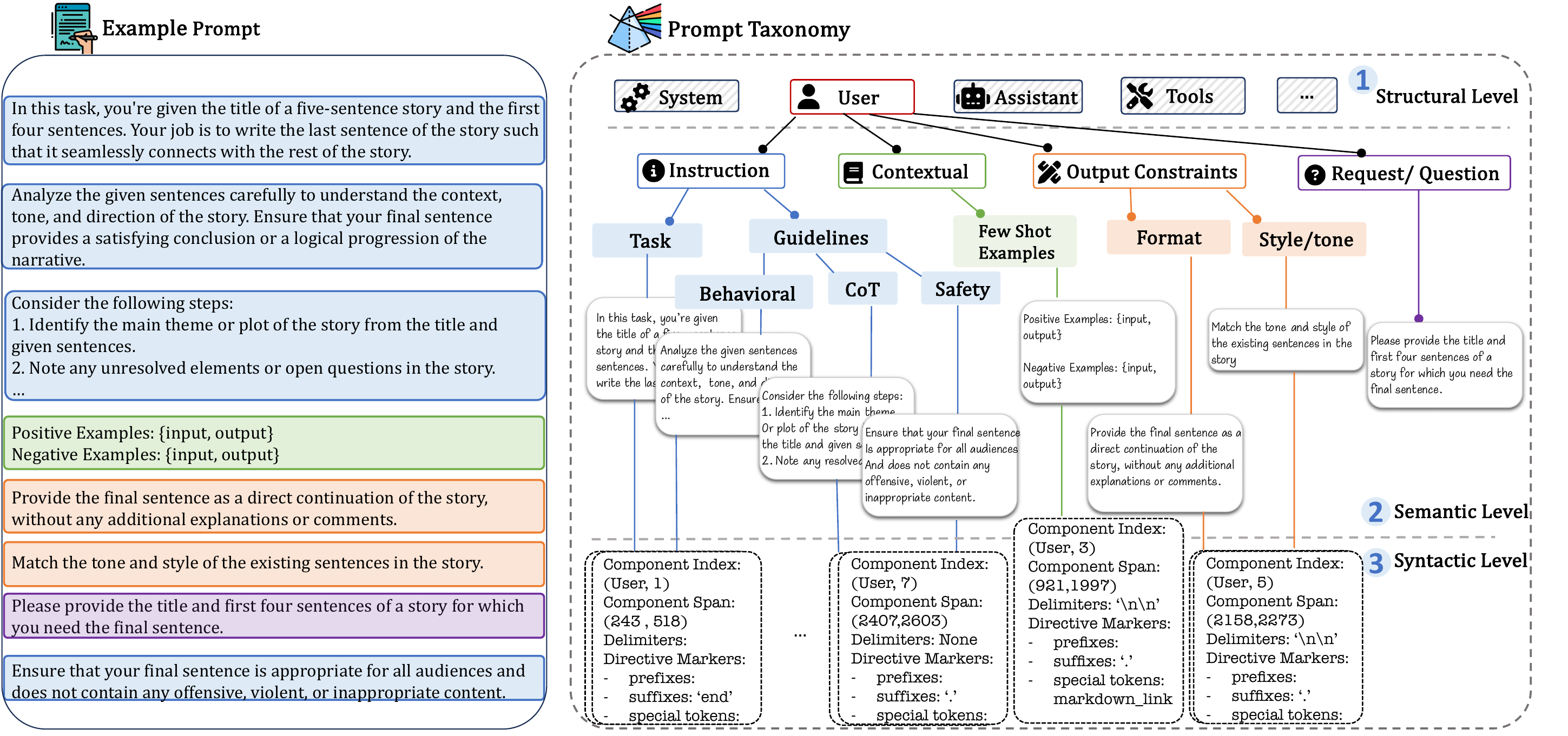}
    \caption{\prismsmall \textsc{PromptPrism} Framework. \small{The \textsc{PromptPrism} taxonomy decomposes an example prompt from Super Natural Instruction dataset (Left) into three hierarchical levels (Right). (1) \textbf{Structural}, corresponding to specified roles in the prompt (e.g. System, User, Assistant) ; (2) \textbf{Semantic}, detailing the content's semantic components (e.g. Instruction, Output Constraints); and (3) \textbf{Syntactic}, describing the syntactic structure of each semantic element (e.g. Delimiters, Special tokens).}}
    \label{fig:promptprism_framework}
\end{figure*}


To address these limitations and meet the demands of increasingly sophisticated LLM applications, we propose a comprehensive and linguistically-inspired taxonomy that integrates structural, semantic, and syntactic dimensions of prompt analysis. Our framework, \prismsmall \textsc{PromptPrism}, treats prompts as structured discourse units with plan-based intentions, enabling hierarchical decomposition and analysis across multiple dimensions. This approach allows for systematic examination of prompt characteristics that influence model behavior, providing insights that purely empirical approaches might miss. 



Our research makes two primary contributions. (1) First, we propose a novel taxonomy framework for prompts that synthesizes linguistic theory with practical applications, enabling systematic analysis and categorization of prompt characteristics (\S \ref{sec:promptprism}). (2) Second, we demonstrate the practical utility of our framework through three distinct applications: taxonomy-guided prompt refinement (\S \ref{sec:prompt-refinement}), comprehensive dataset profile extraction (\S \ref{sec:dataset-profile}), and systematic prompt sensitivity analysis (\S \ref{sec:sensitivity-analysis}). This work represents a significant step toward understanding prompts, enhancing the prompt quality, and enabling controlled experimental setups for LLM development and evaluation.

%% file: 1_related_works.tex
\section{Related Works}
\subsection{Taxonomy of Prompts}

Recent research has introduced various taxonomic approaches for analyzing LLM prompts \cite{bach2022promptsource, schulhoff2024prompt}, with notable contributions including the TELeR framework \cite{santu2023teler}, which categorizes prompts into data and instructions components with difficulty-based classification, and \citet{budagam2024hierarchical}'s hierarchical prompting taxonomy grounded in cognitive principles. While these frameworks have advanced our understanding of prompt structure and evaluation, they remain limited in their ability to capture and differentiate fine-grained components of prompts, such as semantic variations and stylistic subtleties. 


\subsection{Understanding Prompts}
The systematic study of prompt design and optimization has emerged as a critical research area in the evolution of Large Language Models. While effective prompts can significantly enhance model performance \cite{wei2022chain, shinn2023reflexion}, studies have revealed that LLMs demonstrate notable sensitivity to various prompt characteristics, including stylistic formatting \cite{sclarquantifying,he2024does}, sequential ordering \cite{liu2024lost,huang2025math}, task explanations \cite{li2023large}, and meta-rules \cite{li2024think}. This sensitivity, coupled with inconsistent documentation of prompt specifications across the literature, presents challenges for systematic performance evaluation and comparative analysis. This necessitates a structured approach to understand and analyze prompts prompts such as the organization of semantic components and their stylistic representations. 

Due to limited space, the related works regarding structured prompts can be found in Appendix \ref{related_work_structured_prompts}.

%% file: 2_prompt_prism.tex
\section{\textsc{PromptPrism}}\label{sec:promptprism}
We define a prompt $P$ as an ordered sequence of $n$ pairs, where each pair consists of content and its associated role, formally expressed as:
$$P = {(r_i,c_i)}_{ i =1}^n,  r_i \in R$$
where $R = \{\text{system},\text{user}, \text{assistant}, \text{tools} \dots \}$ denotes the finite set of possible roles. The content $c_i$ belongs to a modality space $M$, which encompasses both single modalities and their combinations: $M \subseteq \{\text{text}, \text{image}, \text{audio}\} \cup \{m_1 \otimes m_2 \mid m_1, m_2 \in M\}$ 
Here, $\otimes$ represents the operation of combining different modalities. While our framework accommodates multi-modal prompts through this formalization, the present work focuses on text modality. For instance, an example of text-modality prompt $P$ would be, \texttt{(system, `You`re a helpful assistant')}, where \texttt{system} corresponds to the role $r$, and the \texttt{content} corresponds to the $c$. 

In the development of our framework, we adhere to a top-down, property-guided formation approach. The framework is designed to meet the following key criteria: 
\begin{doubbox}
\noindent \greencheck \ \textbf{Simplicity}: We prioritize minimal complexity in classification, ensuring clear parent-child relationships and intuitive categorization boundaries. This approach facilitates ease of understanding and application across diverse prompt types.

\noindent \greencheck \ \textbf{Coherence}: The framework maintains logical consistency across categories, minimizing redundancy in classification. We establish well-defined relationships between elements, ensuring a cohesive structure that accurately represents the interplay between different prompt components.
\end{doubbox}

\begin{doubbox}
\noindent \greencheck \ \textbf{Comprehensiveness}: Our taxonomy aims for complete coverage of possible prompt types, incorporating both atomic and composite structures. The framework is designed with scalability in mind, allowing for the accommodation of new modalities as the field of prompt evolves.

\noindent \greencheck \ \textbf{Utility}: The framework is developed with practical applicability. It facilitates systematic prompt design and enables meaningful analysis of prompts. This utility-focused approach ensures that the taxonomy serves as a valuable tool for both researchers and practitioners in the field. 
\end{doubbox}

\subsection{Prompt Taxonomy}
The proposed taxonomy adopts a hierarchical tree-like structure, where the structural level decomposes into semantic components, each containing associated syntactic information (shown in Fig \ref{fig:promptprism_framework})

\textcircled{\small{1}} \textbf{Structural Level} At the structural level, \textsc{PromptPrism} conceptualizes prompts as organized discourse units, drawing parallels with traditional linguistic analysis of text structure \cite{mann1992rhetorical,grosz1986attention}. This approach is similar to how linguistics examines the organization of different text genres, such as academic articles with their standard sections (introduction, methodology, results). 

In the context of prompts, this structural organization is defined through roles, $r \in R$, each serving distinct discourse purposes. A key aspect of this structural level is the clear delineation of roles within the prompt, similar to the speaker-listener roles in discourse analysis \cite{santu2023teler,brown2020language}. Specifically, we identify and distinguish between different roles in the prompt interaction: \texttt{System, User, Assistant, Tools}. 
\begin{singlebox}
 - \small{\texttt{System}: Provides core instructions and framework}\\
 - \texttt{User}: Initiates queries or requests\\
 - \texttt{Assistant}: Generates responses and ensures conversation flows naturally\\
 - \texttt{Tools}: Handles specific functionalities or operations
\end{singlebox}

While these four roles form the primary set in our current analysis (Table \ref{tab:roles}), it's important to note that this role system is extensible and can accommodate new roles as prompt architectures evolve across different providers.

\begin{table*}[h]
  \centering
  \resizebox{\textwidth}{!}{%
    \begin{tabular}{lp{20.75em}l}
    \toprule
    \rowcolor[rgb]{ .906,  .902,  .902} \multicolumn{1}{c}{\textbf{\texttt{Semantic Components}}} & \multicolumn{1}{c}{\textbf{\texttt{Description}}} & \multicolumn{1}{c}{\textbf{\texttt{Examples}}} \\
    \midrule
    \midrule
    \rowcolor[rgb]{ .867,  .922,  .969} │\faInfoCircle\  \textbf{\texttt{Instruction}}  & \multicolumn{2}{p{48em}}{\texttt{High-level directive component that guide the model's behavior}}  \\
    \midrule
    \rowcolor[rgb]{ .941,  .941,  .941}\ \ \ \ ├ \texttt{Task}  & \texttt{Task related instruction}  & \multicolumn{1}{p{26.25em}}{This is a classification task.} \\
    \multicolumn{1}{p{15.5em}}{\ \ \ \ ├ \texttt{Guidelines} } & \multicolumn{2}{p{40em}}{\texttt{Non-task specific instructions that shape response behavior}}    \\
    \rowcolor[rgb]{ .941,  .941,  .941}\ \ \ \ \ \ \ \ ├ \texttt{Role assumption (persona)} & Directives for the model to assume a specific role  & \multicolumn{1}{p{26.25em}}{"Suppose you are a \{persona\}.." \cite{perez-etal-2023-discovering} } \\
    \ \ \ \ \ \ \ \ ├ \texttt{Scenario}  & Context setting for the task environment & Imagine we gave you \$2000 right now..\newline{}\cite{binz2023cognitive} \\
    \rowcolor[rgb]{ .941,  .941,  .941}\ \ \ \ \ \ \ \ ├ \texttt{Behavioral}  & Instructions for model conduct and interaction style  & Before you respond, rephrase the question \cite{deng2022rlprompt} \\
    \ \ \ \ \ \ \ \ ├ \texttt{Emotion} & Guidelines for emotional tone  & You'd better be sure \cite{li2023emotionpromp} \\
    \rowcolor[rgb]{ .941,  .941,  .941}\ \ \ \ \ \ \ \ ├ \texttt{Chain of Thought Instruction} & Directives for showing reasoning process & \multicolumn{1}{p{26.25em}}{Let's think step by step \cite{wei2022chain}} \\
    \ \ \ \ \ \ \ \ ├ \texttt{Safety}  & Guidelines ensuring safe and ethical responses  & \multicolumn{1}{p{26.25em}}{"Avoid harmful content", "Maintain ethical boundaries"} \\
    \midrule
    \rowcolor[rgb]{ .867,  .922,  .969} │ \faBook \ \texttt{\textbf{Contextual/Reference info}}  & \multicolumn{2}{p{48em}}{\texttt{Background information and supporting materials}} \\
    \midrule
    \rowcolor[rgb]{ .941,  .941,  .941}\ \ \ \ ├ \texttt{Few-shot examples}     & Sample input-output pairs for in-context learning & \multicolumn{1}{p{26.25em}}{Q: What is the capital of France? A: Paris} \\
    \ \ \ \ ├ \texttt{Knowledge Base}  & Reference information or facts  & \multicolumn{1}{p{26.25em}}{"Based on medical guidelines…"} \\
    \rowcolor[rgb]{ .941,  .941,  .941}\ \ \ \ ├ \texttt{Context for task} & Relevant background information & \multicolumn{1}{p{26.25em}}{ {Text for summarization}} \\
    \midrule
    \rowcolor[rgb]{ .867,  .922,  .969} │\faPencilRuler\ \textbf{\texttt{Output Constraints }} & \multicolumn{2}{p{48em}}{\texttt{Specifications for response format and limitations}}   \\
    \midrule
    \rowcolor[rgb]{ .941,  .941,  .941}\ \ \ \ ├ \texttt{Label Space} & Defined set of possible output categories & \multicolumn{1}{p{26.25em}}{"Choose from: [Positive, Negative, Neutral]"} \\
    \ \ \ \ ├ \texttt{Word Limits}  & Restrictions on response length  & \multicolumn{1}{p{26.25em}}{"Respond in 50 words or less", "Provide a one-paragraph answer"} \\
    \rowcolor[rgb]{ .941,  .941,  .941}\ \ \ \ ├ \texttt{Output format Specification} & Structure requirements for the response & \multicolumn{1}{p{26.25em}}{"Format as JSON", "Present in bullet points"} \\
    \ \ \ \ ├ \texttt{Style/tone}  & Requirements for writing style & \multicolumn{1}{p{26.25em}}{"Use academic language", "Write in conversational tone"} \\
    \midrule
    \rowcolor[rgb]{ .867,  .922,  .969} │\faTools\ \textbf{\texttt{Tools}} & \multicolumn{2}{p{48em}}{\texttt{Specifications for tool usage}}   \\
    \midrule
    \rowcolor[rgb]{ .941,  .941,  .941}\ \ \ \ ├\texttt{Tool name}  & Identifier for specific tool & \multicolumn{1}{p{26.25em}}{"Calculator", "Python Interpreter"} \\
    \ \ \ \ ├\texttt{Tool description}  & Explanation of tool functionality  & \multicolumn{1}{p{26.25em}}{"This tool performs statistical analysis"} \\
    \rowcolor[rgb]{ .941,  .941,  .941}\ \ \ \ ├\texttt{Parameters} & Required inputs and configuration  & \multicolumn{1}{p{26.25em}}{"Input format:(x,y)", Required parameters: date\_range, metric"} \\
    \midrule
    \rowcolor[rgb]{ .867,  .922,  .969} │\faQuestionCircle\ \textbf{\texttt{User request / Question}} & \texttt{The primary query or task from user}  & \multicolumn{1}{p{26.25em}}{"What is the capital of France?" (e.g. question in Q\&A task)} \\
    \midrule
    \rowcolor[rgb]{ .867,  .922,  .969} │\faRobot \ \textbf{\texttt{Response}} & \multicolumn{2}{p{48em}}{\texttt{Semantic component for Model's output}}   \\
    \midrule
    \rowcolor[rgb]{ .941,  .941,  .941}\ \ \ \ ├\texttt{Answer} & Direct Answer from the request / question & \multicolumn{1}{p{26.25em}}{"Paris is the capital of France"} \\
    \ \ \ \ ├\texttt{Peripheral-Explanation}  & Supporting information and clarifications  & \multicolumn{1}{p{26.25em}}{"This city has been the capital since.."} \\
    \midrule
    \rowcolor[rgb]{ .867,  .922,  .969} │\faStickyNote \ \textbf{\texttt{Other purposeful components}} & \multicolumn{2}{p{48em}}{\texttt{Additional functional elements}} \\
    \midrule
    \rowcolor[rgb]{ .941,  .941,  .941}\ \ \ \ ├\texttt{Adversarial} & Components designed for adversarial purpose & \multicolumn{1}{p{26.25em}}{"!!!!!" \cite{zou2023universal}} \\
    \midrule
    \rowcolor[rgb]{ .867,  .922,  .969} │\faRocketchat \ \textbf{\texttt{Historical Context }} & \multicolumn{2}{p{48em}}{\texttt{Pointer to the previous conversation history}}   \\
    \rowcolor[rgb]{ .867,  .922,  .969} │\faCog \ \textbf{\texttt{System Prompt }} & \multicolumn{2}{p{48em}}{\texttt{Pointer to the System Prompt}}    \\
    \rowcolor[rgb]{ .867,  .922,  .969} │\faToolbox \ \textbf{\texttt{Tools Prompt }} & \multicolumn{2}{p{48em}}{\texttt{Pointer to the Tools Prompt}}    \\
    \bottomrule
    \end{tabular}%
    }%
    \caption{Semantic Components defined in \textsc{Promptprism} taxonomy}
  \label{tab:semantic-level}%
\end{table*}%

\textcircled{\small{2}} \textbf{Semantic Level} The semantic layer of \textsc{PromptPrism} establishes a systematic framework for analyzing the meaning and intentional structure of prompts, drawing direct parallels with linguistic pragmatics and semantic theory \cite{levinson1983pragmatics, grice1975logic}. This layer decomposes the content $c_i$ of a structural role $r_i$, into the semantic components $(c_i,r_i)\in P$ that serve specific functions. 


Table \ref{tab:semantic-level} presents the overview of semantic components, with primary semantic components that mirror linguistic functional units \cite{searle1969speech, austin1962things}. \faInfoCircle\ \texttt{Instruction} serve as the core semantic component, articulating the fundamental task or objective, similar to the main predicate in linguistic structures \cite{chomsky2014aspects, fillmore1967case}. These are complemented by \faBook\  \texttt{Contextual/Reference information}, which provides the necessary background knowledge and examples, functioning analogously to contextual presuppositions in linguistic pragmatics \cite{sperber1986relevance, carston2008thoughts}. \faPencilRuler\  \texttt{Output Constraints} define formatting and stylistic expectations, parallel to linguistic register and genre conventions. \faQuestionCircle\ \texttt{User request / Question} refers to the primary query or task from user, such as a question in question and answering task. \faTools\ \texttt{Tools} indicate the specification of tools - tool name, tool description, and parameters. 

These components are defined in terms of their functional purpose rather than abstract linguistic concepts, making them accessible to practitioners while maintaining their grounding in linguistic theory. The semantic relationships between these components exhibit what we term "dominance and satisfaction-precedence relations", wherein lower-level segments contribute incrementally to satisfying higher-level discourse purposes.

\textcircled{\small{3}} \textbf{Syntactic Level} The syntactic level of \textsc{PromptPrism} draws parallels with linguistic morphology while adapting these concepts for the unique requirements of prompts. As previous works \cite{sclarquantifying, he2024does} have demonstrated that syntactic components of prompts such as stylistic patterns and ordering influence the model performance, this level introduces syntactic elements that collectively create a complete structural analysis framework. 
 \underline{\textbf{Component Organization}}: A component organization system plays on two primary mechanisms. (1) Component Position assigns unique identifiers (\texttt{Role, Index}) to each prompt element, similar to syntactic tree structures in linguistics \cite{chomsky2014aspects,tesniere2015elements}. (2) Component Span Analysis defines exact boundaries through position markers (\texttt{start\_pos,end\_pos}), enabling  manipulation and reconstruction of prompt components. This component organization information enables prompt sensitivity analysis, reconstructing and perturbing various components. 

\noindent \textbf{\underline{Directive Markers}}: Our framework adapts linguistic concepts of morphemes for prompt analysis, but with a distinction. While linguistic morphology treats prefixes and suffixes as meaningful units that modify word meanings, we define directive markers purely in terms of their structural and sequential properties \cite{matthews1972inflectional, aronoff1976word}. \underline{Prefix Markers}: Unlike linguistic prefixes that modify word meanings (e.g. `un-' in `unhappy'), our prefix patterns serve as structural indicators: (e.g. \texttt{Hash comments (\#)}, \texttt{Double-slash comments (//)},etc). \underline{Suffix Markers}: Similarly, while linguistic suffixes can change word classes (e.g.`-ness' in `happiness'), our suffix patterns function as structural boundaries: Colon endings \texttt{(:)}, Sentence terminators \texttt{(.!?)}, etc. \underline{Delimiters}: We define the delimiter as the boundary between components. 
Detailed definition can be found in Table \ref{tab:syntactic-components} and in Section \ref{syntactic_def}.


%% file: 3_case_study.tex
\section{Applications of \prismsmall \textsc{PromptPrism} }

\begin{figure}[h]
    \centering
    \includegraphics[width=\textwidth]{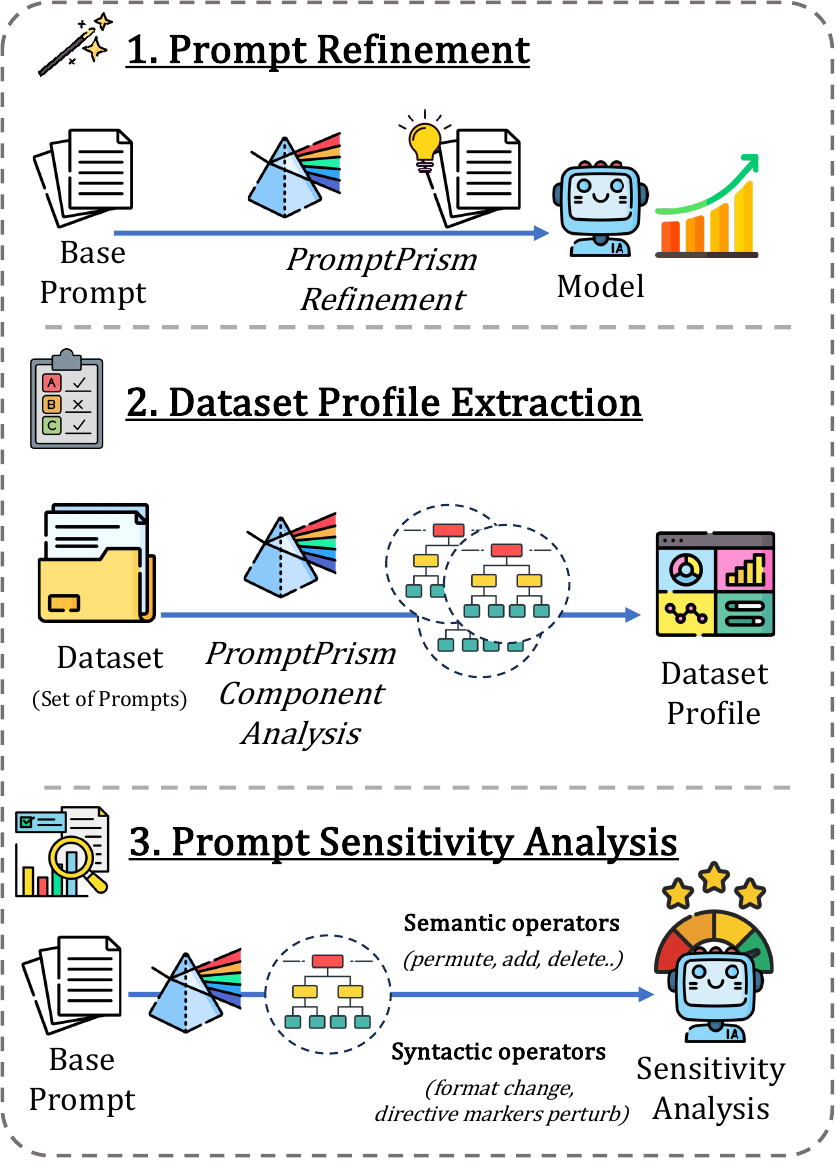}
    \caption{Applications of \textsc{PromptPrism}}
    \label{fig:promptprism-app}
\end{figure}

We demonstrate the applications of \textsc{PromptPrism} and illustrate how it can be effectively leveraged in real-world scenarios. As shown in Figure \ref{fig:promptprism-app}, we focus on three important applications.

\begin{figure*}[h]
    \centering
    \includegraphics[width=\textwidth]{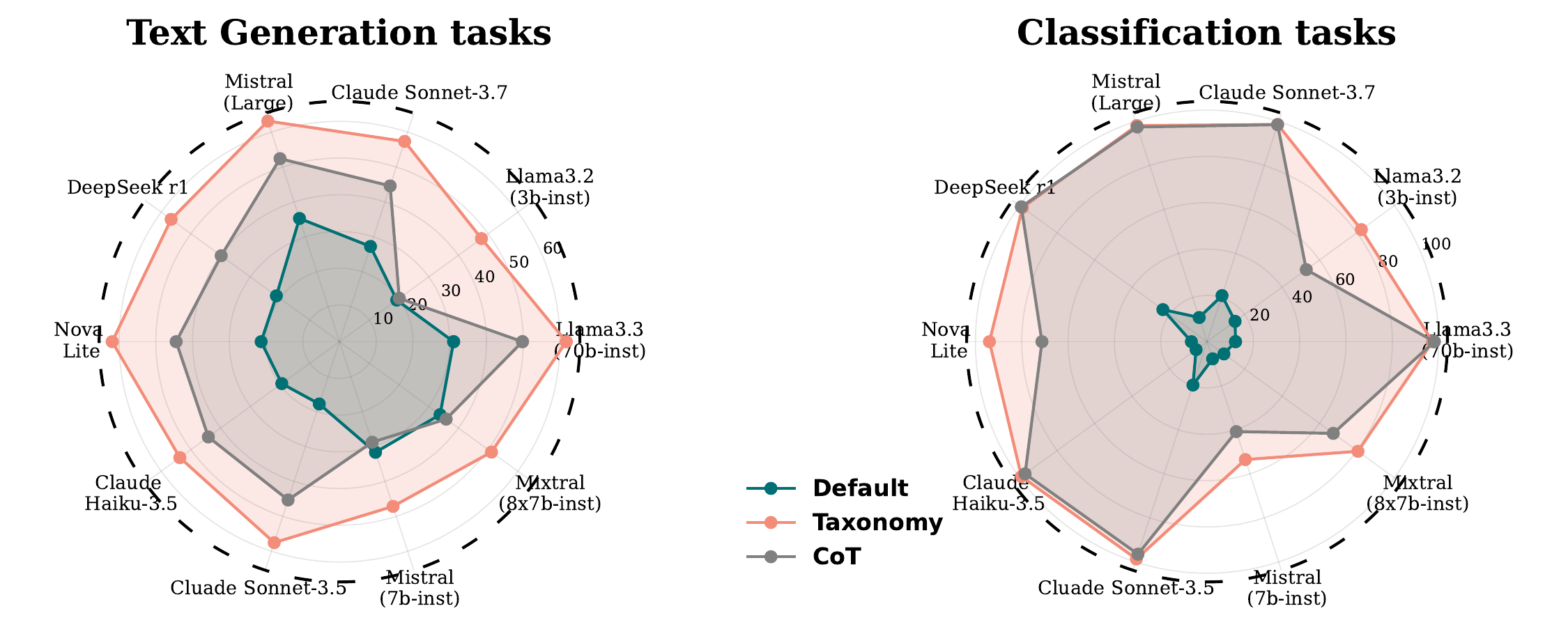}
    \caption{Performance comparison of taxonomy-guided prompt refinement across task types. Our taxonomy-guided approach demonstrates improved performance on text generation tasks (average \textbf{29\%} improvement than CoT) and matches or exceeds CoT baseline on classification tasks (average \textbf{0.13\%} improvement than CoT) . Complete performance metrics are presented in Table \ref{tab:prompt-results-detail}.}
    \label{fig:taxonomy-apo-spider}
\end{figure*}
 
\subsection{Taxonomy Guided Prompt Refinement}\label{sec:prompt-refinement}
We demonstrate how \textsc{PromptPrism} can be used to systematically analyze and improve prompt effectiveness. Our taxonomy guided prompt refinement approach replaces arbitrary prompt elaboration with a more semantically-rich approach, resulting in enhanced model performance compared to the base instructions. 

\noindent \textbf{Experiment Setup} \underline{Dataset}:
We conduct our experiments using Super Natural Instruction dataset \cite{wang2022super} (version 2.8), which provides each task with a base instruction, multiple data instances (consisting of question-answer pairs), and exemplars for few-shot learning. The detailed experiment setting is illustrated in Appendix \ref{sec:prompt_refinement_setting}.

\noindent \underline{Metrics}: For quantitative evaluation, we employ \textsc{Rouge-L} metrics, adhering to the evaluation protocol established in \cite{wang2022super}. The prompt refinement process is conducted using \texttt{Claude-sonnet-3.5} \cite{claude3.5sonnet}, with prompts provided in Appendix \ref{prompt-app}. For each task, we: (1) Generate 5 distinct prompts (setting \texttt{temperature=0.7}), (2) Sample 10 data instances randomly, (3) Perform 50 inference runs per task. We present the result based on using 2 few-shot examples in the main script. The results of 0-shot experiment are presented in the appendix Table \ref{tab:num-few-shot-0}.

\noindent \underline{Baseline} (1) \textbf{Default}: utilizes unmodified base instructions from the Super Natural Instruction dataset \cite{wang2022super}. (2) Chain-of-Thought (\textbf{CoT}) \cite{wei2022chain}: appends the base instruction with \texttt{"Please think step by step and then solve the task."} at the end.\footnote{While some prompt optimization methodologies exist such as DSPy and TextGrad \cite{yuksekgonul2024textgrad, khattab2023dspy}, they require iterative prompt generation and optimization processes, and development set for parameter tuning. To ensure practical applicability of cost-effectiveness, we constrain our experimental setup to single-prompt scenarios, eliminating the need for multiple prompt iterations to achieve optimization without any parameter tuning.} \footnote{We provide additional baseline results of OpenAI Meta prompt approach in Table \ref{tab:openai_baseline}} 

\textbf{Results} Our experimental results, presented in Figure \ref{fig:taxonomy-apo-spider} and tables (Tab \ref{tab:prompt-results-detail}), demonstrate the efficacy of our taxonomy-guided prompt refinement approach across varying configurations. The approach exhibits particularly strong performance in zero-shot settings (Table \ref{tab:num-few-shot-0}), achieving substantial improvements over CoT baselines: a 112\% performance increase in text generation tasks (CoT avg: 40.53, Ours: 55.81) and around 10\% improvement in classification tasks (CoT avg: 81.93, Ours: 90.1).

The benefits of our methodology persist in few-shot scenarios as well. In the two-shot setting (Table \ref{tab:prompt-results-detail}), we observe consistent improvements across task types, with a 29\% enhancement in text generation performance and a modest but consistent 0.13\% improvement in classification tasks. 

\begin{figure*}[h]
    \centering
    \includegraphics[width=\textwidth]{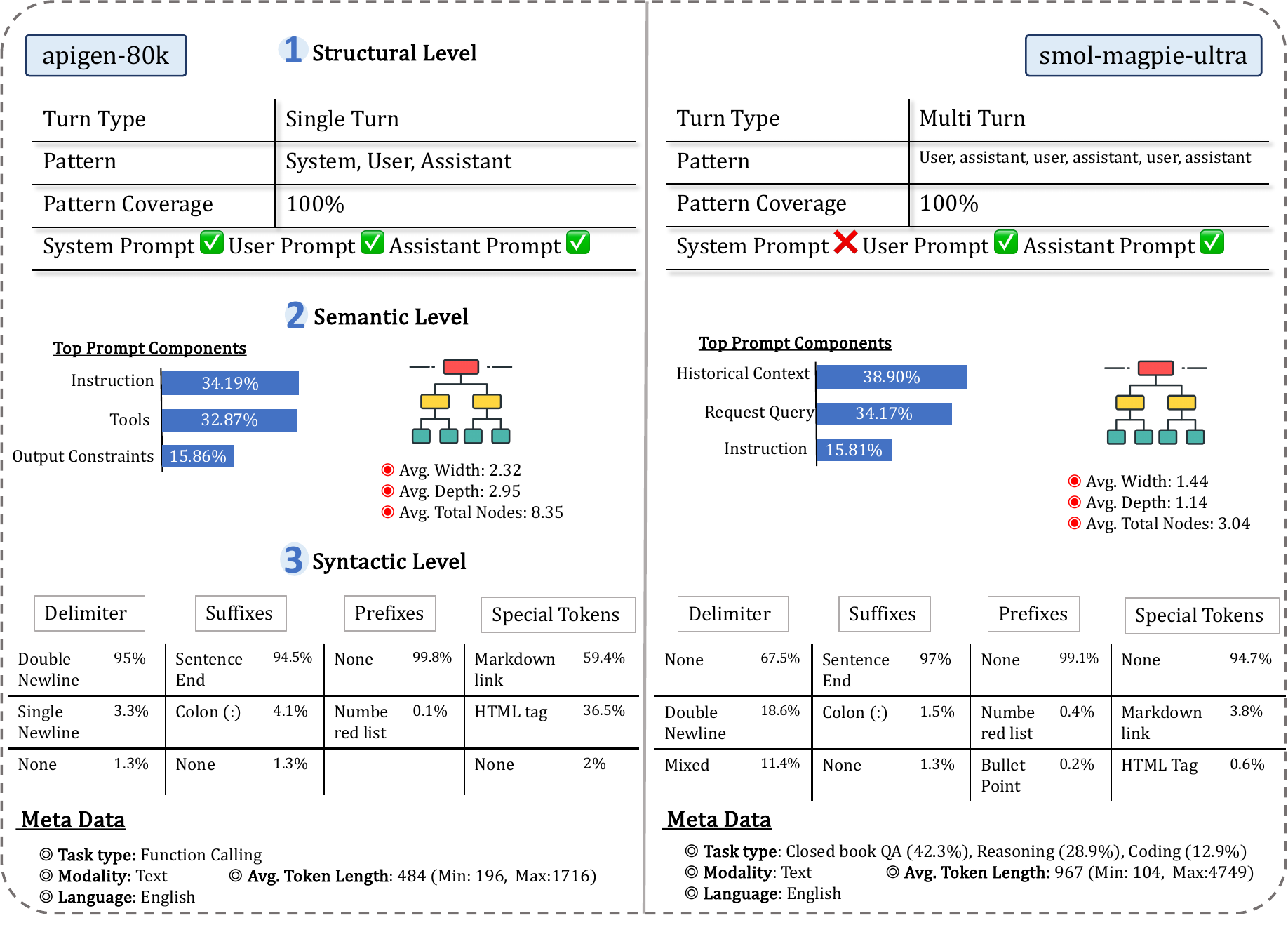}
    \caption{Dataset Profile of \textsc{apigen-80k} \cite{liu2024apigen} and \textsc{smol-magpie-ultra} \cite{allal2025smollm2} on four primary dimensions: (1) Structural Level (2) Semantic Level (3) Syntactic level and (4) Metadata.}
    \label{fig:Dataset_profile}
\end{figure*}

\begin{table*}[h]
  \centering
  \caption{Semantic Component Order Sensitivity Analysis. Performance variations under different component orderings relative to baseline configuration. Asterisks (***) denote statistically significant differences (ANOVA, p < 0.05). Numbers in parentheses represent standard deviations, and percentages indicate relative performance changes from baseline.}
    \resizebox{\textwidth}{!}{%
     \begin{tabular}{ccccccccccc}
\cmidrule{2-11}          & \multicolumn{10}{c}{Ordering Sensitivity Analysis } \\
    \midrule
    \multirow{2}[4]{*}{Model} & \multirow{2}[4]{*}{Baseline} & \multicolumn{3}{c}{Instruction } & \multicolumn{3}{c}{Request/Question} & \multicolumn{3}{c}{Contextual Reference} \\
\cmidrule{3-11}          &       & First & Middle & Last  & First & Middle & Last  & First & Middle & Last \\
    \midrule
    \multirow{3}[2]{*}{Claude-Sonnet-3.5***} & 56.49 & 55.77 & 48.99 & 63.37 & 24.74 & 13.46 & 55.97 & 41.31 & 53.83 & 52.39 \\
          & \textcolor[rgb]{ .682,  .667,  .667}{(0.03)} & \textcolor[rgb]{ .682,  .667,  .667}{(0.04)} & \textcolor[rgb]{ .682,  .667,  .667}{(0.04)} & \textcolor[rgb]{ .682,  .667,  .667}{(0.09)} & \textcolor[rgb]{ .682,  .667,  .667}{(0.02)} & \textcolor[rgb]{ .682,  .667,  .667}{(0.01)} & \textcolor[rgb]{ .682,  .667,  .667}{(0.10)} & \textcolor[rgb]{ .682,  .667,  .667}{(0.06)} & \textcolor[rgb]{ .682,  .667,  .667}{(0.06)} & \textcolor[rgb]{ .682,  .667,  .667}{(0.06)} \\
          &       & \textcolor[rgb]{ .957,  .69,  .518}{\textbf{-1\%}} & \textcolor[rgb]{ .957,  .69,  .518}{\textbf{-13\%}} & \textcolor[rgb]{ .439,  .678,  .278}{\textbf{12\%}} & \textcolor[rgb]{ .957,  .69,  .518}{\textbf{-56\%}} & \textcolor[rgb]{ .957,  .69,  .518}{\textbf{-76\%}} & \textcolor[rgb]{ .957,  .69,  .518}{\textbf{-1\%}} & \textcolor[rgb]{ .957,  .69,  .518}{\textbf{-27\%}} & \textcolor[rgb]{ .957,  .69,  .518}{\textbf{-5\%}} & \textcolor[rgb]{ .957,  .69,  .518}{\textbf{-7\%}} \\
    \midrule
    \multirow{3}[2]{*}{Llama3.2-3b-inst***} & 47.21 & 46.54 & 45.97 & 49.54 & 35.86 & 31.73 & 58.58 & 51.97 & 48.49 & 51.33 \\
          & \textcolor[rgb]{ .682,  .667,  .667}{(0.08)} & \textcolor[rgb]{ .682,  .667,  .667}{(0.06)} & \textcolor[rgb]{ .682,  .667,  .667}{(0.07)} & \textcolor[rgb]{ .682,  .667,  .667}{(0.15)} & \textcolor[rgb]{ .682,  .667,  .667}{(0.04)} & \textcolor[rgb]{ .682,  .667,  .667}{(0.03)} & \textcolor[rgb]{ .682,  .667,  .667}{(0.11)} & \textcolor[rgb]{ .682,  .667,  .667}{(0.09)} & \textcolor[rgb]{ .682,  .667,  .667}{(0.06)} & \textcolor[rgb]{ .682,  .667,  .667}{(0.11)} \\
          &       & \textcolor[rgb]{ .957,  .69,  .518}{\textbf{-1\%}} & \textcolor[rgb]{ .957,  .69,  .518}{\textbf{-3\%}} & \textcolor[rgb]{ .439,  .678,  .278}{\textbf{5\%}} & \textcolor[rgb]{ .957,  .69,  .518}{\textbf{-24\%}} & \textcolor[rgb]{ .957,  .69,  .518}{\textbf{-33\%}} & \textcolor[rgb]{ .439,  .678,  .278}{\textbf{24\%}} & \textcolor[rgb]{ .439,  .678,  .278}{\textbf{10\%}} & \textcolor[rgb]{ .439,  .678,  .278}{\textbf{3\%}} & \textcolor[rgb]{ .439,  .678,  .278}{\textbf{9\%}} \\
    \midrule
    \multirow{3}[2]{*}{Mixtral 8x7b inst ***} & 67.82 & 64.12 & 45.13 & 52.23 & 56.87 & 73.63 & 62.10 & 69.63 & 64.07 & 67.90 \\
          & \textcolor[rgb]{ .682,  .667,  .667}{(0.06)} & \textcolor[rgb]{ .682,  .667,  .667}{(0.06)} & \textcolor[rgb]{ .682,  .667,  .667}{(0.03)} & \textcolor[rgb]{ .682,  .667,  .667}{(0.10)} & \textcolor[rgb]{ .682,  .667,  .667}{(0.03)} & \textcolor[rgb]{ .682,  .667,  .667}{(0.12)} & \textcolor[rgb]{ .682,  .667,  .667}{(0.11)} & \textcolor[rgb]{ .682,  .667,  .667}{(0.25)} & \textcolor[rgb]{ .682,  .667,  .667}{(0.07)} & \textcolor[rgb]{ .682,  .667,  .667}{(0.13)} \\
          &       & \textcolor[rgb]{ .957,  .69,  .518}{\textbf{-5\%}} & \textcolor[rgb]{ .957,  .69,  .518}{\textbf{-33\%}} & \textcolor[rgb]{ .957,  .69,  .518}{\textbf{-23\%}} & \textcolor[rgb]{ .957,  .69,  .518}{\textbf{-16\%}} & \textcolor[rgb]{ .439,  .678,  .278}{\textbf{9\%}} & \textcolor[rgb]{ .957,  .69,  .518}{\textbf{-8\%}} & \textcolor[rgb]{ .439,  .678,  .278}{\textbf{3\%}} & \textcolor[rgb]{ .957,  .69,  .518}{\textbf{-6\%}} & \textcolor[rgb]{ .439,  .678,  .278}{\textbf{0\%}} \\
    \midrule
    \multirow{3}[2]{*}{Deepseek r1} & 67.46 & 67.62 & 54.10 & 61.16 & 62.35 & 59.90 & 63.12 & 58.20 & 65.18 & 66.40 \\
          & \textcolor[rgb]{ .682,  .667,  .667}{(0.14)} & \textcolor[rgb]{ .682,  .667,  .667}{(0.09)} & \textcolor[rgb]{ .682,  .667,  .667}{(0.10)} & \textcolor[rgb]{ .682,  .667,  .667}{(0.16)} & \textcolor[rgb]{ .682,  .667,  .667}{(0.06)} & \textcolor[rgb]{ .682,  .667,  .667}{(0.05)} & \textcolor[rgb]{ .682,  .667,  .667}{(0.20)} & \textcolor[rgb]{ .682,  .667,  .667}{(0.11)} & \textcolor[rgb]{ .682,  .667,  .667}{(0.16)} & \textcolor[rgb]{ .682,  .667,  .667}{(0.06)} \\
          &       & \textcolor[rgb]{ .957,  .69,  .518}{\textbf{0\%}} & \textcolor[rgb]{ .957,  .69,  .518}{\textbf{-20\%}} & \textcolor[rgb]{ .957,  .69,  .518}{\textbf{-9\%}} & \textcolor[rgb]{ .957,  .69,  .518}{\textbf{-8\%}} & \textcolor[rgb]{ .957,  .69,  .518}{\textbf{-11\%}} & \textcolor[rgb]{ .957,  .69,  .518}{\textbf{-6\%}} & \textcolor[rgb]{ .957,  .69,  .518}{\textbf{-14\%}} & \textcolor[rgb]{ .957,  .69,  .518}{\textbf{-3\%}} & \textcolor[rgb]{ .957,  .69,  .518}{\textbf{-2\%}} \\
    \bottomrule
    \end{tabular}%
    }
  \label{tab:ordering-sensitivity}%
\end{table*}%
\subsection{Multi-Dimensional Dataset Profiling}\label{sec:dataset-profile}
\textsc{PromptPrism} enables comprehensive profiling of prompt datasets, extracting and aggregating characteristics across structural, semantic, and syntactic dimensions. This analysis helps identify gaps in existing benchmark datasets and guide the development of more detailed benchmark documentation. 

Our analysis encompasses four primary dimensions:  \underline{(1) Structural Analysis}: We examine fundamental structural characteristics including turn type (single/multi-turn), prompt pattern (role sequence analysis), and unique structural roles (system, user, assistant,etc). 
\underline{(2) Dataset Annotation \& Semantic Analysis}: We conducted systematic annotation of data instances using the \textsc{PromptPrism} taxonomy framework with \texttt{Claude-Sonnet-3.5} \cite{claude3.5sonnet} for semantic component extraction. The annotation process utilizes a structured XML-style format (\texttt{<tag>..<$\backslash$tag>}), enabling efficient extraction and analysis of semantic components (annotation prompt detailed in Appendix \ref{prompt: annotation}). The semantic analysis encompasses two key elements: component frequency analysis, identifying the three most prevalent semantic components in the dataset, and hierarchical tree structure analysis, examining mean taxonomic tree width and mean tree depth to measure component breadth and semantic complexity. 
\underline{(3) Syntactic Analysis}: From annotated data, we extract component-level syntactic information, including component indices, spans, delimiters, and directive markers.
\underline{(4) Metadata Information Analysis}: We derive task type with \texttt{Claude-Sonnet-3.5} using the meta prompt in Appendix \ref{prompt: annotation}. We also include other information such as language specification, token length, and modality characteristics. 

\textbf{Experiment Setup} \underline{Dataset}: Our analysis focuses on two distinct sub-datasets from \textsc{SmolTalk} \cite{allal2025smollm2}: \textsc{apigen-80k} \cite{liu2024apigen} and \textsc{smol-magpie-ultra} \cite{allal2025smollm2}. These datasets were selected for their distinct characteristics: \textsc{apigen-80k} exemplifies function-calling scenarios with tool utilization, while \textsc{smol-magpie-ultra} represents multi-turn interactions. This selection demonstrates \textsc{PromptPrism}'s applicability beyond single-turn interactions to more complex scenarios\footnote{For multi-turn data analysis, we focus on the terminal user prompt, treating preceding interactions as historical context. The detailed prompt configuration can be found in Appendix \ref{prompt-app}; We present human validation analysis in Appendix \ref{human_annotate}.}. 

\textbf{Results} Figure \ref{fig:Dataset_profile} presents a comprehensive profile analysis of the examined datasets across multiple analytical dimensions. \underline{Structural Analysis:} 
\textsc{apigen-80k} implements a single-turn architecture with a consistent system-user-assistant sequence, while \textsc{smol-magpie-ultra} employs a multi-turn structure characterized by alternating user-assistant interactions across three turns. \underline{Semantic Composition:} \textsc{apigen-80k} showed a prominence in instruction, tools, and output constraint components, whereas \textsc{smol-magpie-ultra} emphasizes historical context, request queries, and instructions. Semantic tree structures indicates that \textsc{apigen-80k} exhibits greater component diversity, as evidenced by increased tree width and total node count. This structural richness contrasts with the relatively streamlined composition of \textsc{smol-magpie-ultra}, though it should be noted that our analysis focuses solely on terminal user prompts in multi-turn interactions, potentially affecting these metrics. \underline{Syntactic Characteristics:} we observed that \textsc{apigen-80k} employs consistent formatting conventions, utilizing double newlines as delimiters and incorporating markdown-style special tokens. In contrast, \textsc{smol-magpie-ultra} demonstrates minimal use of explicit delimiters or special tokens, suggesting a more monotone syntactic structure. \underline{Metadata} The \textsc{apigen-80k} consists of monolithic task type, function calling, while the \textsc{smol-magpie-ultra} composes of diverse task such as closed book QA, Reasoning, and Coding etc.

\subsection{Prompt Sensitivity Analysis}\label{sec:sensitivity-analysis}
We leverage \textsc{PromptPrism} to establish a framework for assessing prompt sensitivity. Our approach introduces two classes of operators: semantic operators (permute, add, delete) and syntactic operators (format and delimiter modifications), enabling systematic evaluation of model responses to prompt variations.

\textbf{Experiment Setup} \underline{Data}: We conducted controlled experiments using taxonomy-guided prompts generated in Section \ref{sec:prompt-refinement}. We selected task \texttt{Task067:abductivenli answer generation} as our experimental focus, chosen for its below-average performance score (56.49) relative to the mean text generation task score (73.95). All other experimental parameters remain consistent with those described in Section \ref{sec:prompt-refinement}. \underline{Model}: We conduct experiments with model \texttt{Claude-Sonnet-3.5} and \texttt{Llama3.2-3b-inst}. 

\underline{Semantic operation}: We conduct a systematic investigation of semantic component reordering sensitivity. While the framework supports arbitrary component intervention and positioning, we focus our analysis on three primary semantic components (Instruction, Question, and Few-shot examples). These components are systematically repositioned at the beginning, middle, and end of the prompt sequence to assess performance variations. The detailed implementation methodology is illustrated in Figure \ref{fig:reorder-component}.

\underline{Syntactic operation}: Our syntactic analysis focuses on delimiter modification operations. We examine four distinct delimiter patterns: double newline (`$\backslash$n$\backslash$n'), section separator (`$\backslash$n\#\#\#\#\#$\backslash$n'), tab ($\backslash$t'), and extended whitespace (`$\backslash$s+'). 
Implementation details are provided in Figure \ref{fig:delimiter-mod}.

\textbf{Results} Our analysis of semantic component ordering sensitivity, presented in Table \ref{tab:ordering-sensitivity}, reveals significant effects across multiple models: \texttt{Claude-Sonnet-3.5}, \texttt{Llama3.2-3b-inst}, and \texttt{Mixtral 8x7 inst} demonstrate statistically significant performance variations in response to component reordering. Notably, positioning the Instruction component at the terminal position within the prompt sequence yields the most substantial performance improvements: a 12\% enhancement for \texttt{Claude-Sonnet-3.5} and a 5\% improvement for \texttt{Llama3.2-3b-inst}. Interestingly, the Deepseek-r1 model showed more robustness to component order variations, with no statistically significant differences observed.

In contrast, our investigation of delimiter sensitivity (Table \ref{tab:Delimiter-sensitivity}) indicates that although there are differences in model performances, modifications to delimiter patterns do not produce statistically significant performance variations. This suggests that while models exhibit marked sensitivity to semantic ordering, they maintain robustness to syntactic delimiter variations.

%% file: 4_conclusion.tex
\section{Conclusion}
In this paper, we present \prismsmall\textsc{PromptPrism}, a comprehensive taxonomy framework for prompt analysis grounded in linguistic principles. Our framework provides a systematic approach to understanding and analyzing prompt composition, structure, and effectiveness across various applications. This work contributes toward standardizing prompt analysis in the field of LLMs. 


\section*{Limitations}
While \textsc{PromptPrism} provides a systematic framework for prompt analysis, several limitations warrant acknowledgment. First, our approach employs a top-down methodology in establishing the taxonomy. Although grounded in linguistic principles, this approach may not capture all emergent patterns that could be identified through bottom-up analysis of existing prompt collections. We acknowledge that no single taxonomic framework can claim to be definitive or completely comprehensive in the rapidly evolving field of prompt engineering.

Furthermore, the current implementation lacks extensive manual annotation of existing large-scale datasets. While our framework demonstrates effectiveness in automated analysis,and present pilot human validation in Section \ref{human_annotate}, 
the absence of human-validated annotations across diverse prompt collections limits our ability to fully validate the taxonomy's generalizability and robustness across different domains and applications.

These limitations suggest valuable directions for future work, including: Integration of bottom-up analysis methods to complement our top-down approach,
Development of comprehensive manually annotated benchmark datasets

\section*{Broader Impact}
The development and application of this work carry significant implications for the field of natural language processing and artificial intelligence at large. This work adheres to the ACL Code of Ethics and is primarily aimed at enhancing the transparency and interpretability of language models. 

By providing a systematic framework for analyzing and understanding prompts, \textsc{PromptPrism} has the potential to: 

\textbf{Enhancing Transparency and Reproducibility}
\textsc{PromptPrism} addresses a critical challenge in current LLM research: the inconsistent documentation of prompts in publications, which hinders reproducibility and comparative analysis. Just as Model Cards \cite{mitchell2019model} standardized the documentation of model characteristics, \textsc{PromptPrism} provides a structured framework for documenting prompt characteristics. This standardization enables more precise reporting of prompt components in research papers, facilitating more meaningful comparisons across studies and enhancing the reproducibility of results.

\textbf{Practical Applications Beyond Research}
Beyond academic research, \textsc{PromptPrism} offers practical value for organizations developing LLM applications. The framework can help standardize prompt libraries and documentation within organizations, facilitating knowledge transfer between prompt engineers and enabling more systematic prompt management. This standardization is particularly valuable as organizations scale their LLM applications and need to maintain consistency across multiple systems and use cases.





While the immediate risks associated with this research are low, we acknowledge that improved prompt engineering techniques could potentially be misused to create more persuasive or manipulative AI generated content. However, we believe that the benefits of increased transparency and interpretability outweigh these potential risks.

%% file: 5_appendix.tex
\newpage{}
\section*{Appendix}
\label{sec:appendix}
\section{Prompt Refinement Experiment Setting}\label{sec:prompt_refinement_setting}
\textbf{Data} We focused on a diverse set of 70 tasks spanning two primary categories: 53 classification tasks and 17 text generation tasks, following \cite{sclarquantifying}. 

17 tasks selected for text generation: \texttt{task037, task038, task040, task067, task071, task072, task105, task216, task223, task240, task348, task389, task443, task845, task1326, task1401,  task1613}. 

53 Tasks used for classification: \texttt{task050, task065, task069, task070, task114, task133, task155, task158, task161, task162, task163, task190, task213, task214, task220, task279, task280, task286, task296, task297, task316, task317, task319, task320, task322, task323, task325, task326, task327, task328, task335, task337, task385, task580, task607, task608, task609, task904, task905, task1186, task1283, task1284, task1297, task1347, task1387, task1419, task1420, task1421, task1423, task1502, task1612, task1678, task1724.}

\noindent \textbf{Models}: The models for evaluation include Claude 3.7 Sonnet \cite{claude3.7sonnet}, Claude 3.5 Sonnet, Haiku \cite{claude3.5sonnet,claude3.5haiku}, Llama3.2-3b-inst \cite{llama3.2-3b-inst}, Llama3.3-70b-inst \cite{llama3.3-70b-inst}, Mistral-7b-inst \cite{mistral7b-inst}, Mistral Large \cite{mistrallarge}, Mixtral8x7b-inst \cite{mixtral8x7b-inst}, Nova Lite \cite{novalite}, DeepSeek-r1 \cite{deepseekr1}. For inference, we set the \texttt{temperature=0}, for reproducibility.

\section{Syntactic Definition}\label{syntactic_def}
Table \ref{tab:syntactic-components} and Figure \ref{fig:syntactic-impl} shows the detailed definition of syntactic components and implementation details. Figure \ref{fig:delimiter-analysis} shows snippet of code analyzing delimiters.

\section{Annotation Validation Analysis}\label{human_annotate}
\begin{table}[htbp]
  \centering
  \resizebox{\textwidth}{!}{%
  \caption{Human annotator agreement analysis. Mean annotation scores (0-1 scale) across criteria. Values in parentheses show Pearson correlation coefficients (p) between annotators, where 1.0 indicates perfect agreement.}
 \label{tab:human_annotate}%
    \begin{tabular}{lccc}
    \toprule
    Dataset & Format correctness & Tag correctness & Coverage  \\
    \midrule
    apigen-80k & 0.99 (1.0) & 1.0 (1.0) & 0.98 (0.98) \\
    smol-magpie-ultra  & 1.0 (1.0) & 0.97 (0.79) & 0.92 (0.55) \\
    \bottomrule
    \end{tabular}%
    }
\end{table}%
To validate the LLM-based annotation described in Section \ref{sec:dataset-profile}, we conducted an evaluation using two independent human verifiers across three key dimensions: 

\noindent \textbf{Format Correctness} evaluates the structural integrity of XML-style annotations, specifically the proper matching of \texttt{<tag>..<$\backslash$tag>} pairs. We quantify correctness as the ratio of properly formatted tags to total tags, normalized to a scale of 0 to 1, where 1 indicates perfect format conformity. 

\noindent \textbf{Tag Correctness} assesses the semantic accuracy of content classification within tags. This metric evaluates whether content is appropriately categorized (e.g., whether text marked as `instruction' truly represents instructional content). Scores are normalized on 0-1 scale, with 1 representing complete semantic accuracy. 

\noindent \textbf{Coverage} measures the comprehensiveness of content annotation, evaluating whether all relevant content is appropriately tagged. For instance, if a multi-sentence instruction is only partially tagged, the coverage score is reduced accordingly. Each component receives a binary score (0 or 1), based on complete coverage of semantic units. 

Analysis of the evaluation results, presented in Table \ref{tab:human_annotate}, demonstrates strong performance across all three assessment metrics—format correctness, tag correctness, and coverage—for both datasets. The metrics exhibit high mean values (exceeding 0.95), indicating robust annotation quality. Inter-annotator agreement shows strong consistency, with correlation coefficients ranging from 0.55 to 1.0, suggesting reliable assessment methodology. These results indicate that our LLM-based annotation approach achieves high accuracy and reproducibility across different annotators.

\begin{table*}[h]
  \centering
  \resizebox{\textwidth}{!}{%
    \begin{tabular}{p{10.085em}p{17.585em}p{12.665em}}
    \toprule
    \rowcolor[rgb]{ .867,  .922,  .969} \multicolumn{1}{c}{\textbf{\texttt{Syntactic Components}}} & \textbf{\texttt{Description}} & \textbf{\texttt{Examples}} \\
    \midrule
    \midrule
    │Component Indexing & Maps semantic component to role and order & (User, 0) \# User, order index \\
    \rowcolor[rgb]{ .941,  .941,  .941} │Component Span & Spans of a component  & (start\_position, end\_position) \newline{}\# absolute values  \\
    │Delimiters  & Boundary between components  & Double newline ("$\backslash$n$\backslash$n"), Single newline ("$\backslash$n"), Tab ("$\backslash$t"), Whitespace ("$\backslash$s+") \\
    \rowcolor[rgb]{ .941,  .941,  .941} │Directive Markers  & \multicolumn{2}{p{30.25em}}{Special tokens or pattens that signal component structure} \\
    \ \ \ \ ├Prefixes  & Markers at the beginning of components & hash comments (\#) , double-slash comments (//),  blockquotes (>),  numbered lists (1.,2.,etc),  and bullet points (-,*,+) \\
    \rowcolor[rgb]{ .941,  .941,  .941}\ \ \ \ ├Suffixes  & Markers at the end of components & Colon(:), Sentence End (.!?), Semicolon (;) \\
    \ \ \ \ ├Special token  & Special tokens specific to models  & Llama3.1:<|begin\_of\_text|>, <|end\_of\_text|>, <|eom\_id|>\newline{}Mistral-7b-inst: <s>..</s>, [INST]..[/INST] \\
        \bottomrule
    \end{tabular}%
    }
    \caption{Syntactic Components Definition}
  \label{tab:syntactic-components}%
\end{table*}%

\begin{figure*}[h]
\begin{minted}
[frame=lines,
framesep=2mm,
baselinestretch=1.2,
bgcolor=LightGray,
fontsize=\footnotesize,
linenos]
{python}
# prefix patterns
prefix_patterns = [
    (r"^\s*#[^#\n]+", "hash_comment"),
    (r"^\s*//[^\n]+", "double_slash_comment"),
    (r"^\s*>[^\n]+", "blockquote"),
    (r"^\s*\d+\.\s", "numbered_list"),
    (r"^\s*[-*+]\s", "bullet_point"),
]

# suffix patterns
suffix_patterns = [
    (r"\s*:\s*$", "colon_end"),
    (r"\s*[.!?]+$", "sentence_end"),
    (r"\s*[;]\s*$", "semicolon_end"),
]

# Special token patterns
special_token_patterns = [
    (r"<[^>]+>", "html_tag"),
    (r"\[.*?\]", "markdown_link"),
    (r"\$.*?\$", "math_expression"),
    (r"@\w+", "mention"),
    (r"#\w+", "hashtag"),
    (r"https?://\S+", "url"),
]
\end{minted}
\caption{Syntactic Components Implementation}
\label{fig:syntactic-impl}
\end{figure*}

\begin{figure*}[h]
\begin{minted}
[frame=lines,
framesep=2mm,
%baselinestretch=1.2,
bgcolor=LightGray,
fontsize=\footnotesize,
%linenos
]
{python}
def _analyze_delimiter(self, delimiter_text):
        """Analyze the delimiter between components"""
        if delimiter_text is None:
            return None

        delimiter_info = {"raw": delimiter_text, "length": len(delimiter_text), "type": None, "pattern": None}

        # Analyze delimiter pattern
        if delimiter_text.isspace():
            if "\n\n" in delimiter_text:
                delimiter_info["type"] = "double_newline"
                delimiter_info["pattern"] = r"\n\n"
            elif "\n" in delimiter_text:
                delimiter_info["type"] = "single_newline"
                delimiter_info["pattern"] = r"\n"
            elif "\t" in delimiter_text:
                delimiter_info["type"] = "tab"
                delimiter_info["pattern"] = r"\t"
            else:
                delimiter_info["type"] = "whitespace"
                delimiter_info["pattern"] = r"\s+"
        else:
            delimiter_info["type"] = "mixed"
            delimiter_info["pattern"] = repr(delimiter_text)[1:-1]
        return delimiter_info
\end{minted}
\caption{Syntactic-Delimiter analyzing implementation}
\label{fig:delimiter-analysis}
\end{figure*}

\begin{table*}[ht]
  \centering
  \resizebox{\textwidth}{!}{%
    \begin{tabular}{cp{13.415em}p{8.585em}p{9.835em}p{10.585em}}
    \toprule
    \rowcolor[rgb]{ .906,  .902,  .902} \textbf{Providers} & \textbf{System} & \textbf{User} & \textbf{Assistant} & \textbf{Tools} \\
    \midrule
    \multirow{2}{*}{\parbox{7.165em}{\centering Bedrock \\ Converse API}} & \cellcolor[rgb]{ .941,  .941,  .941}\bluecheck & \cellcolor[rgb]{ .941,  .941,  .941}\bluecheck & \cellcolor[rgb]{ .941,  .941,  .941}\bluecheck & \cellcolor[rgb]{ .941,  .941,  .941}\bluecheck \\
    & (Implicit, i.e., 'System' is not stated in message)\newline{}\newline{}prompt that provides instructions or context to the model about the task it should perform, or the persona it should adopt during the conversation. & The human that is sending messages to the model. & The model that is sending messages back to the human user. & The definition of the tool is a JSON schema that you pass in the toolConfig (ToolConfiguration) request parameter to the Converse operation \\
    \midrule
    \multirow{2}{*}{OpenAI} & \cellcolor[rgb]{ .941,  .941,  .941}\bluecheck $\blacktriangleright$ (Developer) & \cellcolor[rgb]{ .941,  .941,  .941}\bluecheck & \cellcolor[rgb]{ .941,  .941,  .941}\bluecheck & \cellcolor[rgb]{ .941,  .941,  .941}\bluecheck \\
    & Instructions to the model that are prioritized ahead of user messages, following chain of command. Previously called the system prompt. & Input from end users, or a catch-all for data we want to provide to the model & Sampled from the language model & Generated by some program, such as code execution or an API call \\
    \midrule
    \multirow{2}{*}{Anthropic \newline{} Claude} & \cellcolor[rgb]{ .941,  .941,  .941}\bluecheck & \cellcolor[rgb]{ .941,  .941,  .941}\bluecheck & \cellcolor[rgb]{ .941,  .941,  .941}\bluecheck & \cellcolor[rgb]{ .941,  .941,  .941}\bluecheck \\
    & (Implicit, i.e., 'System' is not stated in message)\newline{}\newline{}A system prompt is a way of providing context and instructions to Claude, such as specifying a particular goal or role. & & & \\
    \midrule
    \multirow{2}{*}{Llama} & \cellcolor[rgb]{ .941,  .941,  .941}\bluecheck & \cellcolor[rgb]{ .941,  .941,  .941}\bluecheck & \cellcolor[rgb]{ .941,  .941,  .941}\bluecheck & \cellcolor[rgb]{ .941,  .941,  .941}\bluecheck $\blacktriangleright$ (ipython) \\
    & Sets the context in which to interact with the AI model. It typically includes rules, guidelines, or necessary information that help the model respond effectively. & Represents the human interacting with the model. It includes the inputs, commands, and questions to the model. & Represents the response generated by the AI model based on the context provided in the system, ipython and userprompts. & A new role introduced in Llama 3.1. Semantically, this role means "tool". This role is used to mark messages with the output of a tool call when sent back to the model from the executor. \\
    \midrule
    \multirow{2}{*}{Microsoft Azure} & \cellcolor[rgb]{ .941,  .941,  .941}\bluecheck $\blacktriangleright$(Developer) for GPT-o family & \cellcolor[rgb]{ .941,  .941,  .941}\bluecheck & \cellcolor[rgb]{ .941,  .941,  .941}\bluecheck & \cellcolor[rgb]{ .941,  .941,  .941}\bluecheck \\
    & A brief description of the assistant.\newline{}Personality traits of the assistant.\newline{}Instructions or rules you want the assistant to follow.\newline{}Data or information needed for the model, such as relevant questions from an FAQ. & End user message & Assistant is a large language model trained by OpenAI. & A list of tools the model may call. Currently, only functions are supported as a tool. Use this to provide a list of functions the model may generate JSON inputs for. A max of 128 functions are supported. \\
    \midrule
    Mistral API & \cellcolor[rgb]{ .941,  .941,  .941}\bluecheck & \cellcolor[rgb]{ .941,  .941,  .941}\bluecheck & \cellcolor[rgb]{ .941,  .941,  .941}\bluecheck & \cellcolor[rgb]{ .941,  .941,  .941}\bluecheck \\
    \midrule
    \multirow{2}{*}{Gemini API} & \cellcolor[rgb]{ .941,  .941,  .941}\bluecheck $\blacktriangleright$(systemInstruction) & \cellcolor[rgb]{ .941,  .941,  .941}\bluecheck & \cellcolor[rgb]{ .941,  .941,  .941}\bluecheck $\blacktriangleright$ (model) & \cellcolor[rgb]{ .941,  .941,  .941}\bluecheck \\
    & Optional. Developer set system instruction(s). Currently, text only. & Available in chat mode & Available in chat mode & Optional. A list of Tools the Model may use to generate the next response.\newline{}A Tool is a piece of code that enables the system to interact with external systems to perform an action, or set of actions, outside of knowledge and scope of the Model. Supported Tools are Function and codeExecution. \\
    \bottomrule
    \end{tabular}%
  }
  \caption{Structural Roles Across LLM Platforms. Comparison of role definitions across major LLM service providers. While platforms generally support four standard roles (\texttt{System, User, Assistant, Tools}), some employ platform-specific nomenclature. Our framework adopts these standardized role definitions to ensure broad applicability and consistency.}
  \label{tab:roles}
\end{table*}

\begin{table*}[h]
  \centering
  \caption{Taxonomy Guided Prompt Refinement Results in 2-shot setting. Performance improvements over Chain-of-Thought (CoT) baseline using our taxonomy-guided approach. Text generation tasks show an average 29\% improvement across models, while classification tasks demonstrate a 0.13\% enhancement. Percentages indicate relative performance gains compared to CoT and the numbers in parenthesis indicate standard deviation.}
  \resizebox{\textwidth}{!}{%
    \begin{tabular}{cccc|ccc}
    \toprule
    \cmidrule{2-7}          & \multicolumn{6}{c}{Number of Few Shots: 2 } \\
    \midrule
    \multirow{2}[4]{*}{Model } & \multicolumn{3}{c}{\texttt{Text Generation Tasks }} & \multicolumn{3}{c}{\texttt{Classification Tasks}} \\
\cmidrule{2-7}          & \textsc{Default} & \textsc{CoT}   & \textsc{Taxonomy} & \textsc{Default} & \textsc{CoT}   & \textsc{Taxonomy} \\
    \midrule
    \multirow{3}[1]{*}{Claude Sonnet 3.7 } & 27.28 & 44.6  & \textbf{57.35} & 20.92 & 98.55 & 98.68 \\
          & \textcolor[rgb]{ .647,  .647,  .647}{(0.2)} & \textcolor[rgb]{ .647,  .647,  .647}{(0.22)} & \textcolor[rgb]{ .647,  .647,  .647}{(0.26)} & \textcolor[rgb]{ .647,  .647,  .647}{(0.38)} & \textcolor[rgb]{ .647,  .647,  .647}{(0.11)} & \textcolor[rgb]{ .647,  .647,  .647}{(0.10)} \\
          &       &       & \textcolor[rgb]{ .439,  .678,  .278}{\textbf{+ 29.0} \%} &       &       & \textcolor[rgb]{ .439,  .678,  .278}{\textbf{+ 0.13 \%}} \\
          \hdashline
    \multirow{3}[0]{*}{Claude Sonnet 3.5} & 17.84 & 45.32 & \textbf{57.55} & 19.7  & 96.5  & \textbf{98.68} \\
          & \textcolor[rgb]{ .647,  .647,  .647}{(0.09)} & \textcolor[rgb]{ .647,  .647,  .647}{(0.20)} & \textcolor[rgb]{ .647,  .647,  .647}{(0.23)} & \textcolor[rgb]{ .647,  .647,  .647}{(0.36)} & \textcolor[rgb]{ .647,  .647,  .647}{(0.18)} & \textcolor[rgb]{ .647,  .647,  .647}{(0.10)} \\
          &       &       & \textcolor[rgb]{ .439,  .678,  .278}{\textbf{+ 27 \%}} &       &       & \textcolor[rgb]{ .439,  .678,  .278}{\textbf{+ 2.2 \%}} \\
          \hdashline
    \multirow{3}[0]{*}{Claude Haiku 3.5} & 19.42 & 44.14 & \textbf{53.73} & 5.84  & 97.11 & \textbf{99.06} \\
          & \textcolor[rgb]{ .647,  .647,  .647}{(0.17)} & {\textcolor[rgb]{ .647,  .647,  .647}{ (0.21)}} & {\textcolor[rgb]{ .647,  .647,  .647}{(0.26)}} & {\textcolor[rgb]{ .647,  .647,  .647}{(0.17)}} & \textcolor[rgb]{ .647,  .647,  .647}{(0.16)} & \textcolor[rgb]{ .647,  .647,  .647}{(0.07)} \\
          &       &       & \textcolor[rgb]{ .439,  .678,  .278}{\textbf{+ 21.7 \%}} &       &       & \textcolor[rgb]{ .439,  .678,  .278}{\textbf{+ 2.01 \%}} \\
          \hdashline
    \multirow{3}[0]{*}{Llama3.2 3b inst } & 19.26 & 20.1  & \textbf{47.76} & 14.97 & 52.96 & \textbf{82.35} \\
          & \textcolor[rgb]{ .647,  .647,  .647}{(0.16)} & \textcolor[rgb]{ .647,  .647,  .647}{(0.20)} & \textcolor[rgb]{ .647,  .647,  .647}{(0.24)} & \textcolor[rgb]{ .647,  .647,  .647}{(0.30)} & \textcolor[rgb]{ .647,  .647,  .647}{(0.48)} & \textcolor[rgb]{ .647,  .647,  .647}{(0.35)} \\
          &       &       & \textcolor[rgb]{ .439,  .678,  .278}{\textbf{+ 137.6 \%}} &       &       & \textcolor[rgb]{ .439,  .678,  .278}{\textbf{+ 55.5 \%}} \\
          \hdashline
    \multirow{3}[0]{*}{Llama3.3 70b inst } & 31.06 & 49.8  & \textbf{61.72} & 12.29 & \textbf{98.11} & 97.39 \\
          & \textcolor[rgb]{ .647,  .647,  .647}{(0.22)} & \textcolor[rgb]{ .647,  .647,  .647}{(0.25)} & \textcolor[rgb]{ .647,  .647,  .647}{(0.26)} & {\textcolor[rgb]{ .647,  .647,  .647}{(0.28)}} & \textcolor[rgb]{ .647,  .647,  .647}{(0.13)} & \textcolor[rgb]{ .647,  .647,  .647}{(0.15)} \\
          &       &       & \textcolor[rgb]{ .439,  .678,  .278}{\textbf{+ 23.9 \%}} &       &       & \textcolor[rgb]{ .973,  .796,  .678}{\textbf{- 0.73 \%}} \\
          \hdashline
    \multirow{3}[0]{*}{Mistral 7b inst} & 31.68 & 28.78 & \textbf{47.16} & 7.79  & 40.84 & \textbf{53.55} \\
          & \textcolor[rgb]{ .647,  .647,  .647}{(0.18)} & \textcolor[rgb]{ .647,  .647,  .647}{(0.21)} & \textcolor[rgb]{ .647,  .647,  .647}{(0.17)} & \textcolor[rgb]{ .647,  .647,  .647}{(0.05)} & \textcolor[rgb]{ .647,  .647,  .647}{(0.44)} & \textcolor[rgb]{ .647,  .647,  .647}{(0.43)} \\
          &       &       & \textcolor[rgb]{ .439,  .678,  .278}{\textbf{+ 68.9 \%}} &       &       & \textcolor[rgb]{ .439,  .678,  .278}{\textbf{+ 31.1 \%}} \\
          \hdashline
    \multirow{3}[0]{*}{Mistral Large} & 35.28 & 52.41 & \textbf{63.17} & 10.98 & 97.48 & \textbf{98.15} \\
          & \textcolor[rgb]{ .647,  .647,  .647}{(0.18)} & \textcolor[rgb]{ .647,  .647,  .647}{(0.23)} & \textcolor[rgb]{ .647,  .647,  .647}{(0.23)} & \textcolor[rgb]{ .647,  .647,  .647}{(0.22)} & \textcolor[rgb]{ .647,  .647,  .647}{(0.13)} & \textcolor[rgb]{ .647,  .647,  .647}{(0.13)} \\
          &       &       & \textcolor[rgb]{ .439,  .678,  .278}{\textbf{+ 22.4 \%}} &       &       & \textcolor[rgb]{ .439,  .678,  .278}{\textbf{+ 0.68 \%}} \\
          \hdashline
    \multirow{3}[0]{*}{Mixtral 8x7b inst} & 33.83 & 35.88 & \textbf{51.14} & 9.00  & 67.37 & \textbf{80.54} \\
          & \textcolor[rgb]{ .647,  .647,  .647}{(0.17)} & \textcolor[rgb]{ .647,  .647,  .647}{(0.21)} & \textcolor[rgb]{ .647,  .647,  .647}{(0.19)} & \textcolor[rgb]{ .647,  .647,  .647}{(0.13)} & \textcolor[rgb]{ .647,  .647,  .647}{(0.44)} & \textcolor[rgb]{ .647,  .647,  .647}{(0.37)} \\
          &       &       & \textcolor[rgb]{ .439,  .678,  .278}{\textbf{+ 42.5 \%}} &       &       & \textcolor[rgb]{ .439,  .678,  .278}{\textbf{+ 19.5 \%}} \\
          \hdashline
    \multirow{3}[0]{*}{Nova Lite} & 21.34 & 44.47 & \textbf{61.89} & 6.76  & 71.27 & \textbf{93.96} \\
          & \textcolor[rgb]{ .647,  .647,  .647}{(0.08)} & \textcolor[rgb]{ .647,  .647,  .647}{(0.23)} & \textcolor[rgb]{ .647,  .647,  .647}{(0.21)} & \textcolor[rgb]{ .647,  .647,  .647}{(0.03)} & \textcolor[rgb]{ .647,  .647,  .647}{(0.34)} & \textcolor[rgb]{ .647,  .647,  .647}{(0.20)} \\
          &       &       & \textcolor[rgb]{ .439,  .678,  .278}{\textbf{+ 39.2 \%}} &       &       & \textcolor[rgb]{ .439,  .678,  .278}{\textbf{+ 31.8 \%}} \\
          \hdashline
    \multirow{3}[1]{*}{Deepseek r1} & 21.25 & 39.82 & \textbf{56.67} & 23.57 & \textbf{99.12} & 98.5 \\
          & \textcolor[rgb]{ .647,  .647,  .647}{(0.11)} & \textcolor[rgb]{ .647,  .647,  .647}{(0.28)} & \textcolor[rgb]{ .647,  .647,  .647}{(0.26)} & \textcolor[rgb]{ .647,  .647,  .647}{(0.36)} & \textcolor[rgb]{ .647,  .647,  .647}{(0.09)} & \textcolor[rgb]{ .647,  .647,  .647}{(0.09)} \\
          &       &       & \textcolor[rgb]{ .439,  .678,  .278}{\textbf{+ 42.3 \%}} &       &       & \textcolor[rgb]{ .973,  .796,  .678}{\textbf{- 0.63 \%}} \\
    \bottomrule
    \end{tabular}%
    }%
  \label{tab:prompt-results-detail}%
\end{table*}%

\begin{table*}[h]
  \centering
  \caption{Taxonomy Guided Prompt Refinement Results in 0-shot setting. Performance improvements over Chain-of-Thought (CoT) baseline using our taxonomy-guided approach. Text generation tasks show an average 112\% improvement across models, while classification tasks demonstrate a 461\% enhancement. Percentages indicate relative performance gains compared to CoT and the numbers in parenthesis indicate standard deviation.}
  \resizebox{\textwidth}{!}{%
    \begin{tabular}{cccc|ccc}
    \toprule
    \cmidrule{2-7}          & \multicolumn{6}{c}{Number of Few Shots: 0 } \\
    \midrule
    \multirow{2}[4]{*}{Model } & \multicolumn{3}{c}{\texttt{Text Generation Tasks} } & \multicolumn{3}{c}{\texttt{Classification Tasks}} \\
\cmidrule{2-7}          & \textsc{Default} & \textsc{CoT}   & \textsc{Taxonomy} & \textsc{Default} & \textsc{CoT}   & \textsc{Taxonomy} \\
\midrule

    \multirow{3}[0]{*}{Claude Sonnet 3.7 } & 37.62 & 38.75 & \textbf{51.16} & 96.48 & 96.73 & 99.02 \\
          & \textcolor[rgb]{ .647,  .647,  .647}{(0.26)} & \textcolor[rgb]{ .647,  .647,  .647}{(0.25)} & \textcolor[rgb]{ .647,  .647,  .647}{(0.25)} & \textcolor[rgb]{ .647,  .647,  .647}{(0.15)} & \textcolor[rgb]{ .647,  .647,  .647}{(0.15)} & \textcolor[rgb]{ .647,  .647,  .647}{(0.07)} \\
          &       &       & \textcolor[rgb]{ .439,  .678,  .278}{\textbf{+36\%}} &       &       & \textcolor[rgb]{ .439,  .678,  .278}{\textbf{+3\%}} \\
          \hdashline
    \multirow{3}[0]{*}{Claude Sonnet 3.5} & 14.9  & 41.29 & \textbf{50.58} & 11.44 & 94.81 & 98.86 \\
          & \textcolor[rgb]{ .647,  .647,  .647}{(0.06)} & \textcolor[rgb]{ .647,  .647,  .647}{(0.23)} & \textcolor[rgb]{ .647,  .647,  .647}{(0.20)} & \textcolor[rgb]{ .647,  .647,  .647}{(0.26)} & \textcolor[rgb]{ .647,  .647,  .647}{(0.19)} & \textcolor[rgb]{ .647,  .647,  .647}{(0.08)} \\
          &       &       & \textcolor[rgb]{ .439,  .678,  .278}{\textbf{+239\%}} &       &       & \textcolor[rgb]{ .439,  .678,  .278}{\textbf{+764\%}} \\
          \hdashline
    \multirow{3}[0]{*}{Claude Haiku 3.5} & 15.83 & 36.22 & \textbf{45.92} & 5.87  & 95.05 & \textbf{99.05} \\
          & \textcolor[rgb]{ .647,  .647,  .647}{(0.08)} & {\textcolor[rgb]{ .647,  .647,  .647}{(0.23)}} & {\textcolor[rgb]{ .647,  .647,  .647}{(0.21)}} & {\textcolor[rgb]{ .647,  .647,  .647}{(0.16)}} & \textcolor[rgb]{ .647,  .647,  .647}{(0.06)} & \textcolor[rgb]{ .647,  .647,  .647}{(0.07)} \\
          &       &       & \textcolor[rgb]{ .439,  .678,  .278}{\textbf{+190\%}} &       &       & \textcolor[rgb]{ .439,  .678,  .278}{\textbf{+1587\%}} \\
          \hdashline
    \multirow{3}[0]{*}{Llama3.2 3b inst } & 29.52 & 20.44 & \textbf{38.57} & 64.49 & 58.29 & \textbf{81.29} \\
          & \textcolor[rgb]{ .647,  .647,  .647}{(0.24)} & \textcolor[rgb]{ .647,  .647,  .647}{(0.19)} & \textcolor[rgb]{ .647,  .647,  .647}{(0.17)} & \textcolor[rgb]{ .647,  .647,  .647}{(0.46)} & \textcolor[rgb]{ .647,  .647,  .647}{(0.47)} & \textcolor[rgb]{ .647,  .647,  .647}{(0.17)} \\
          &       &       & \textcolor[rgb]{ .439,  .678,  .278}{\textbf{31\%}} &       &       & \textcolor[rgb]{ .439,  .678,  .278}{\textbf{26\%}} \\
          \hdashline
    \multirow{3}[0]{*}{Llama3.3 70b inst } & 33.11 & 36.85 & \textbf{49.94} & 39.59 & 94.52 & \textbf{98.02} \\
          & \textcolor[rgb]{ .647,  .647,  .647}{(0.26)} & \textcolor[rgb]{ .647,  .647,  .647}{(0.25)} & \textcolor[rgb]{ .647,  .647,  .647}{(0.22)} & \textcolor[rgb]{ .647,  .647,  .647}{(0.45)} & \textcolor[rgb]{ .647,  .647,  .647}{(0.22)} & \textcolor[rgb]{ .647,  .647,  .647}{(0.14)} \\
          &       &       & \textcolor[rgb]{ .439,  .678,  .278}{\textbf{+51\%}} &       &       & \textcolor[rgb]{ .439,  .678,  .278}{\textbf{+148\%}} \\
          \hdashline
    \multirow{3}[0]{*}{Mistral 7b inst} & 32.21 & 23.08 & \textbf{44.16} & 5.98  & 52.93 & \textbf{53.08} \\
          & \textcolor[rgb]{ .647,  .647,  .647}{(0.17)} & \textcolor[rgb]{ .647,  .647,  .647}{(0.23)} & \textcolor[rgb]{ .647,  .647,  .647}{(0.19)} & \textcolor[rgb]{ .647,  .647,  .647}{(0.05)} & \textcolor[rgb]{ .647,  .647,  .647}{(0.45)} & \textcolor[rgb]{ .647,  .647,  .647}{(0.44)} \\
          &       &       & \textcolor[rgb]{ .439,  .678,  .278}{\textbf{+37\%}} &       &       & \textcolor[rgb]{ .439,  .678,  .278}{\textbf{+788\%}} \\
          \hdashline
    \multirow{3}[0]{*}{Mistral Large} & 36.65 & 44.06 & \textbf{55.22} & 41.01 & 92.49 & \textbf{98.79} \\
          & \textcolor[rgb]{ .647,  .647,  .647}{(0.23)} & \textcolor[rgb]{ .647,  .647,  .647}{(0.24)} & \textcolor[rgb]{ .647,  .647,  .647}{(0.23)} & \textcolor[rgb]{ .647,  .647,  .647}{(0.45)} & \textcolor[rgb]{ .647,  .647,  .647}{(0.23)} & \textcolor[rgb]{ .647,  .647,  .647}{(0.09)} \\
          &       &       & \textcolor[rgb]{ .439,  .678,  .278}{\textbf{+51\%}} &       &       & \textcolor[rgb]{ .439,  .678,  .278}{\textbf{+141\%}} \\
          \hdashline
    \multirow{3}[0]{*}{Mixtral 8x7b inst} & 28.88 & 32.55 & \textbf{45.05} & 18.14 & 68.04 & \textbf{78.77} \\
          & \textcolor[rgb]{ .647,  .647,  .647}{(0.15)} & \textcolor[rgb]{ .647,  .647,  .647}{(0.20)} & \textcolor[rgb]{ .647,  .647,  .647}{(0.18)} & \textcolor[rgb]{ .647,  .647,  .647}{(0.31)} & \textcolor[rgb]{ .647,  .647,  .647}{(0.41)} & \textcolor[rgb]{ .647,  .647,  .647}{(0.37)} \\
          &       &       & \textcolor[rgb]{ .439,  .678,  .278}{\textbf{+56\%}} &       &       & \textcolor[rgb]{ .439,  .678,  .278}{\textbf{+334\%}} \\
          \hdashline
    \multirow{3}[0]{*}{Nova Lite} & 26.92 & 36.04 & \textbf{54.29} & 12.49 & 59.48 & 94.97 \\
          & \textcolor[rgb]{ .647,  .647,  .647}{(0.17)} & \textcolor[rgb]{ .647,  .647,  .647}{(0.21)} & \textcolor[rgb]{ .647,  .647,  .647}{(0.21)} & \textcolor[rgb]{ .647,  .647,  .647}{(0.27)} & \textcolor[rgb]{ .647,  .647,  .647}{(0.40)} & \textcolor[rgb]{ .647,  .647,  .647}{(0.17)} \\
          &       &       & \textcolor[rgb]{ .439,  .678,  .278}{\textbf{+102\%}} &       &       & \textcolor[rgb]{ .439,  .678,  .278}{\textbf{+660\%}} \\
          \hdashline
    \multirow{3}[0]{*}{Deepseek r1} & 11.28 & 40.01 & \textbf{47.91} & 37.61 & 96.61 & 99.05 \\
          & \textcolor[rgb]{ .647,  .647,  .647}{(0.17)} & \textcolor[rgb]{ .647,  .647,  .647}{(0.25)} & \textcolor[rgb]{ .647,  .647,  .647}{(0.24)} & \textcolor[rgb]{ .647,  .647,  .647}{(0.43)} & \textcolor[rgb]{ .647,  .647,  .647}{(0.18)} & \textcolor[rgb]{ .647,  .647,  .647}{(0.06)} \\
          &       &       & \textcolor[rgb]{ .439,  .678,  .278}{\textbf{+325\%}} &       &       & \textcolor[rgb]{ .439,  .678,  .278}{\textbf{+163\%}} \\
          \bottomrule
    \end{tabular}%
    }
  \label{tab:num-few-shot-0}%
\end{table*}%

\begin{table*}[h]
  \centering
  \caption{\small{Additional \cite{openaimetaprompt} baseline. Using Claude Sonnet 3.5 in zero-shot settings (Tab \ref{tab:num-few-shot-0}), we evaluated performance using prompts rewritten according to OpenAI's guidelines. The results demonstrate that while the OpenAI meta prompt approach outperforms the Chain of Thought baseline, our taxonomy-based approach achieves higher performance in text generation tasks.}}
  \resizebox{\textwidth}{!}{%
    \begin{tabular}{cccc|cccc}
    \toprule
    \multicolumn{4}{c}{Text Generation Tasks } & \multicolumn{4}{c}{Classification Tasks} \\
    \midrule
    Default & CoT   & OpenAI & Taxonomy & Default & CoT   & OpenAI & Taxonomy \\
    \midrule
    14.9  & 41.29 & 42.28 & \textbf{50.58} & 11.44 & 94.81 & 86.79 & \textbf{98.86} \\
    \textcolor[rgb]{ .647,  .647,  .647}{(0.06)} & \textcolor[rgb]{ .647,  .647,  .647}{(0.23)} & \textcolor[rgb]{ .647,  .647,  .647}{(0.245)} & \textcolor[rgb]{ .647,  .647,  .647}{(0.20)} & \textcolor[rgb]{ .647,  .647,  .647}{(0.26)} & \textcolor[rgb]{ .647,  .647,  .647}{(0.19)} & \textcolor[rgb]{ .647,  .647,  .647}{(0.34)} & \textcolor[rgb]{ .647,  .647,  .647}{(0.08)} \\
    \bottomrule
    \end{tabular}%
    }
  \label{tab:openai_baseline}%
\end{table*}%

\begin{table*}[h]
  \centering
  \caption{Syntactic Profile. Comprehensive syntactic information of entire \textsc{Smol-Talk} dataset \cite{allal2025smollm2}, presented in Section \ref{sec:dataset-profile}}
  \resizebox{\textwidth}{!}{%
    \begin{tabular}{cccc|ccc|ccc|ccc}
    \toprule
    \rowcolor[rgb]{ .867,  .922,  .969} \textbf{Dataset } & \multicolumn{3}{c|}{\textbf{Delimiter}} & \multicolumn{3}{c|}{\textbf{Suffixes}} & \multicolumn{3}{c|}{\textbf{Prefixes}} & \multicolumn{3}{c}{\textbf{Special Tokens}} \\
    \midrule
    \multirow{2}[2]{*}{apigen-80k } & Double newline ('\textbackslash{}n\textbackslash{}n') & Single newline ('\textbackslash{}n') & None  & Full stop (.) & Colon (:) & None  & None  & Numbered list & Hash comment & Markdown link & Html tag & None \\
          & \textcolor[rgb]{ .459,  .443,  .443}{0.95} & \textcolor[rgb]{ .459,  .443,  .443}{0.033} & \textcolor[rgb]{ .459,  .443,  .443}{0.013} & \textcolor[rgb]{ .459,  .443,  .443}{0.945} & \textcolor[rgb]{ .459,  .443,  .443}{0.041} & \textcolor[rgb]{ .459,  .443,  .443}{0.013} & \textcolor[rgb]{ .459,  .443,  .443}{0.998} & \textcolor[rgb]{ .459,  .443,  .443}{0.001} & \textcolor[rgb]{ .459,  .443,  .443}{0} & \textcolor[rgb]{ .459,  .443,  .443}{0.594} & \textcolor[rgb]{ .459,  .443,  .443}{0.365} & \textcolor[rgb]{ .459,  .443,  .443}{0.02} \\
    \midrule
    \multicolumn{1}{c}{\multirow{2}[2]{*}{everday-conversations}} & None  & Whitespace & mixed & \multicolumn{3}{c|}{Full stop (.)} & \multicolumn{3}{c|}{None} & \multicolumn{3}{c}{None} \\
          & \textcolor[rgb]{ .459,  .443,  .443}{0.983} & \textcolor[rgb]{ .459,  .443,  .443}{0.017} & \textcolor[rgb]{ .459,  .443,  .443}{0} & \multicolumn{3}{c|}{\textcolor[rgb]{ .459,  .443,  .443}{1}} & \multicolumn{3}{c|}{\textcolor[rgb]{ .459,  .443,  .443}{1}} & \multicolumn{3}{c}{\textcolor[rgb]{ .459,  .443,  .443}{1}} \\
    \midrule
    \multicolumn{1}{c}{\multirow{2}[2]{*}{explore-instruct-rewriting }} & Double newline ('\textbackslash{}n\textbackslash{}n') & Whitespace & Single newline ('\textbackslash{}n') & Full stop (.) & Colon (:) & None  & None  & Bullet point & Hash comment & None  & Markdown link & Html tag \\
          & \textcolor[rgb]{ .459,  .443,  .443}{0.655} & \textcolor[rgb]{ .459,  .443,  .443}{0.216} & \textcolor[rgb]{ .459,  .443,  .443}{0.126} & \textcolor[rgb]{ .459,  .443,  .443}{0.617} & \textcolor[rgb]{ .459,  .443,  .443}{0.382} & \textcolor[rgb]{ .459,  .443,  .443}{0} & \textcolor[rgb]{ .459,  .443,  .443}{0.995} & \textcolor[rgb]{ .459,  .443,  .443}{0.005} & \textcolor[rgb]{ .459,  .443,  .443}{0} & \textcolor[rgb]{ .459,  .443,  .443}{0.993} & \textcolor[rgb]{ .459,  .443,  .443}{0.002} & \textcolor[rgb]{ .459,  .443,  .443}{0.001} \\
    \midrule
    \multirow{2}[2]{*}{longalign} & None  & Double newline & mixed & None  & Full stop (.) & Colon (:)  & None  & Numbered list & Hash comment & None  & Markdown link & Html tag \\
          & \textcolor[rgb]{ .459,  .443,  .443}{0.69} & \textcolor[rgb]{ .459,  .443,  .443}{0.225} & \textcolor[rgb]{ .459,  .443,  .443}{0.08} & \textcolor[rgb]{ .459,  .443,  .443}{0.68} & \textcolor[rgb]{ .459,  .443,  .443}{0.27} & \textcolor[rgb]{ .459,  .443,  .443}{0.04} & \textcolor[rgb]{ .459,  .443,  .443}{0.92} & \textcolor[rgb]{ .459,  .443,  .443}{0.04} & \textcolor[rgb]{ .459,  .443,  .443}{0.02} & \textcolor[rgb]{ .459,  .443,  .443}{0.84} & \textcolor[rgb]{ .459,  .443,  .443}{0.077} & \textcolor[rgb]{ .459,  .443,  .443}{0.035} \\
    \midrule
    \multicolumn{1}{c}{\multirow{2}[2]{*}{metamathqa-50k}} & None  & Double newline & Whitespace & Full stop (.) & None  & Colon (:)  & None  & Numbered list & Hash comment & None  & Markdown link & Mention \\
          & \textcolor[rgb]{ .459,  .443,  .443}{0.643} & \textcolor[rgb]{ .459,  .443,  .443}{0.222} & \textcolor[rgb]{ .459,  .443,  .443}{0.111} & \textcolor[rgb]{ .459,  .443,  .443}{0.952} & \textcolor[rgb]{ .459,  .443,  .443}{0.042} & \textcolor[rgb]{ .459,  .443,  .443}{0.004} & \textcolor[rgb]{ .459,  .443,  .443}{0.998} & \textcolor[rgb]{ .459,  .443,  .443}{0.001} & \textcolor[rgb]{ .459,  .443,  .443}{0} & \textcolor[rgb]{ .459,  .443,  .443}{0.964} & \textcolor[rgb]{ .459,  .443,  .443}{0.036} & \textcolor[rgb]{ .459,  .443,  .443}{0} \\
    \midrule
    \multirow{2}[2]{*}{numia-cot-100k} & None  & Double newline & Whitespace & Full stop (.) & None  & Colon (:)  & None  & Bullet point & Numbered list & None  & Markdown link & Html tag \\
          & \textcolor[rgb]{ .459,  .443,  .443}{0.604} & \textcolor[rgb]{ .459,  .443,  .443}{0.244} & \textcolor[rgb]{ .459,  .443,  .443}{0.091} & \textcolor[rgb]{ .459,  .443,  .443}{0.799} & \textcolor[rgb]{ .459,  .443,  .443}{0.169} & \textcolor[rgb]{ .459,  .443,  .443}{0.028} & \textcolor[rgb]{ .459,  .443,  .443}{0.989} & \textcolor[rgb]{ .459,  .443,  .443}{0.006} & \textcolor[rgb]{ .459,  .443,  .443}{0.004} & \textcolor[rgb]{ .459,  .443,  .443}{0.904} & \textcolor[rgb]{ .459,  .443,  .443}{0.08} & \textcolor[rgb]{ .459,  .443,  .443}{0.014} \\
    \midrule
    \multicolumn{1}{c}{\multirow{2}[2]{*}{openhermes-100k}} & Double newline ('\textbackslash{}n\textbackslash{}n') & None  & Whitespace & Full stop (.) & Colon (:) & None  & None  & Numbered list & Bullet point & None  & Markdown link & Html tag \\
          & \textcolor[rgb]{ .459,  .443,  .443}{0.423} & \textcolor[rgb]{ .459,  .443,  .443}{0.313} & \textcolor[rgb]{ .459,  .443,  .443}{0.215} & \textcolor[rgb]{ .459,  .443,  .443}{0.89} & \textcolor[rgb]{ .459,  .443,  .443}{0.077} & \textcolor[rgb]{ .459,  .443,  .443}{0.027} & \textcolor[rgb]{ .459,  .443,  .443}{0.99} & \textcolor[rgb]{ .459,  .443,  .443}{0.005} & \textcolor[rgb]{ .459,  .443,  .443}{0.003} & \textcolor[rgb]{ .459,  .443,  .443}{0.881} & \textcolor[rgb]{ .459,  .443,  .443}{0.072} & \textcolor[rgb]{ .459,  .443,  .443}{0.026} \\
    \midrule
    \multicolumn{1}{c}{\multirow{2}[2]{*}{self-oss-instruct}} & Double newline ('\textbackslash{}n\textbackslash{}n') & Whitespace & None  & Full stop (.) & Colon (:) & None  & None  & Bullet point & Numbered list & None  & Markdown link & Html tag \\
          & \textcolor[rgb]{ .459,  .443,  .443}{0.557} & \textcolor[rgb]{ .459,  .443,  .443}{0.407} & \textcolor[rgb]{ .459,  .443,  .443}{0.017} & \textcolor[rgb]{ .459,  .443,  .443}{0.942} & \textcolor[rgb]{ .459,  .443,  .443}{0.053} & \textcolor[rgb]{ .459,  .443,  .443}{0.003} & \textcolor[rgb]{ .459,  .443,  .443}{0.969} & \textcolor[rgb]{ .459,  .443,  .443}{0.025} & \textcolor[rgb]{ .459,  .443,  .443}{0.005} & \textcolor[rgb]{ .459,  .443,  .443}{0.724} & \textcolor[rgb]{ .459,  .443,  .443}{0.252} & \textcolor[rgb]{ .459,  .443,  .443}{0.014} \\
    \midrule
    \multicolumn{1}{c}{\multirow{2}[2]{*}{smol-constraints}} & Whitespace & Double newline & mixed & Full stop (.) & Colon (:) & None  & None  & Bullet point & Hash comment & Markdown link & None  & Html tag \\
          & \textcolor[rgb]{ .459,  .443,  .443}{0.581} & \textcolor[rgb]{ .459,  .443,  .443}{0.289} & \textcolor[rgb]{ .459,  .443,  .443}{0.122} & \textcolor[rgb]{ .459,  .443,  .443}{0.983} & \textcolor[rgb]{ .459,  .443,  .443}{0.014} & \textcolor[rgb]{ .459,  .443,  .443}{0.002} & \textcolor[rgb]{ .459,  .443,  .443}{0.966} & \textcolor[rgb]{ .459,  .443,  .443}{0.033} & \textcolor[rgb]{ .459,  .443,  .443}{0.0001} & \textcolor[rgb]{ .459,  .443,  .443}{0.438} & \textcolor[rgb]{ .459,  .443,  .443}{0.43} & \textcolor[rgb]{ .459,  .443,  .443}{0.1302} \\
    \midrule
    \multicolumn{1}{c}{\multirow{2}[2]{*}{smol-magpie-ultra}} & None  & Double newline & Whitespace & Full stop (.) & None  & Colon (:)  & None  & Numbered list & Bullet point & None  & Markdown link & Html tag \\
          & \textcolor[rgb]{ .459,  .443,  .443}{0.675} & \textcolor[rgb]{ .459,  .443,  .443}{0.186} & \textcolor[rgb]{ .459,  .443,  .443}{0.114} & \textcolor[rgb]{ .459,  .443,  .443}{0.97} & \textcolor[rgb]{ .459,  .443,  .443}{0.015} & \textcolor[rgb]{ .459,  .443,  .443}{0.013} & \textcolor[rgb]{ .459,  .443,  .443}{0.991} & \textcolor[rgb]{ .459,  .443,  .443}{0.004} & \textcolor[rgb]{ .459,  .443,  .443}{0.002} & \textcolor[rgb]{ .459,  .443,  .443}{0.947} & \textcolor[rgb]{ .459,  .443,  .443}{0.038} & \textcolor[rgb]{ .459,  .443,  .443}{0.006} \\
    \midrule
    \multicolumn{1}{c}{\multirow{2}[2]{*}{smol-rewrite}} & Double newline ('\textbackslash{}n\textbackslash{}n') & Whitespace & Single newline ('\textbackslash{}n') & \multicolumn{3}{c|}{Full stop (.)} & \multicolumn{3}{c|}{None} & None  & Hash tag & Markdown link \\
          & \textcolor[rgb]{ .459,  .443,  .443}{0.556} & \textcolor[rgb]{ .459,  .443,  .443}{0.436} & \textcolor[rgb]{ .459,  .443,  .443}{0.006} & \multicolumn{3}{c|}{\textcolor[rgb]{ .459,  .443,  .443}{1}} & \multicolumn{3}{c|}{\textcolor[rgb]{ .459,  .443,  .443}{1}} & \textcolor[rgb]{ .459,  .443,  .443}{0.68} & \textcolor[rgb]{ .459,  .443,  .443}{0.255} & \textcolor[rgb]{ .459,  .443,  .443}{0.031} \\
    \midrule
    \multicolumn{1}{c}{\multirow{2}[2]{*}{smol-summarize}} & Double newline ('\textbackslash{}n\textbackslash{}n') & Whitespace & mixed & Full stop (.) & None  & Colon (:)  & \multicolumn{3}{c|}{None} & None  & Markdown link & Mention \\
          & \textcolor[rgb]{ .459,  .443,  .443}{0.697} & \textcolor[rgb]{ .459,  .443,  .443}{0.245} & \textcolor[rgb]{ .459,  .443,  .443}{0.055} & \textcolor[rgb]{ .459,  .443,  .443}{0.914} & \textcolor[rgb]{ .459,  .443,  .443}{0.085} & \textcolor[rgb]{ .459,  .443,  .443}{0} & \multicolumn{3}{c|}{\textcolor[rgb]{ .459,  .443,  .443}{1}} & \textcolor[rgb]{ .459,  .443,  .443}{0.961} & \textcolor[rgb]{ .459,  .443,  .443}{0.021} & \textcolor[rgb]{ .459,  .443,  .443}{0.006} \\
    \midrule
    \multicolumn{1}{c}{\multirow{2}[2]{*}{systemchat-30k}} & Double newline ('\textbackslash{}n\textbackslash{}n') & Single newline ('\textbackslash{}n') & Whitespace & Full stop (.) & None  & Colon (:)  & \multicolumn{3}{c|}{None} & None  & Html tag & Mention \\
          & \textcolor[rgb]{ .459,  .443,  .443}{0.648} & \textcolor[rgb]{ .459,  .443,  .443}{0.273} & \textcolor[rgb]{ .459,  .443,  .443}{0.072} & \textcolor[rgb]{ .459,  .443,  .443}{0.995} & \textcolor[rgb]{ .459,  .443,  .443}{0.003} & \textcolor[rgb]{ .459,  .443,  .443}{0.0009} & \multicolumn{3}{c|}{\textcolor[rgb]{ .459,  .443,  .443}{1}} & \textcolor[rgb]{ .459,  .443,  .443}{0.993} & \textcolor[rgb]{ .459,  .443,  .443}{0.002} & \textcolor[rgb]{ .459,  .443,  .443}{0.002} \\
    \bottomrule
    \end{tabular}%
      }%
  \label{tab:syntactic-analysis}%
\end{table*}%

\begin{table*}[h]
  \centering
  \caption{Delimiter Sensitivity Analysis. Numbers in parentheses represent standard deviations, and percentages indicate relative performance changes from baseline. In contrast to Semantic components re-ordering analysis, delimiter sensitivity do not show statistically significant difference between variations.}
  \Large 
  \resizebox{\textwidth}{!}{%
    \begin{tabular}{cccccc}
    \toprule
    \large Model & \multicolumn{1}{p{8.915em}}{\texttt{`\textbackslash{}n\textbackslash{}n'} \newline{}(double new line) } & \multicolumn{1}{p{5.335em}}{\texttt{`\textbackslash{}n\#\#\#\#\#\textbackslash{}n'} \newline{}(number \#)} & \multicolumn{1}{p{6.665em}}{\texttt{`\textbackslash{}n'} \newline{}(single new line)} & \multicolumn{1}{p{4.165em}}{\texttt{`\textbackslash{}t'} \newline{}(tab)} & \multicolumn{1}{p{5em}}{\texttt{`\textbackslash{}s+'} \newline{}(whitespace)} \\
    \midrule
    \multicolumn{1}{c}{\multirow{3}[2]{*}{\large Claude-Sonnet-3.5}} & 56.49 & 51.34 & 52.25 & 57.07 & 54.5 \\
          & \textcolor[rgb]{ .647,  .647,  .647}{(0.03)} & \textcolor[rgb]{ .647,  .647,  .647}{(0.02)} & \textcolor[rgb]{ .647,  .647,  .647}{(0.02)} & \textcolor[rgb]{ .647,  .647,  .647}{(0.07)} & \textcolor[rgb]{ .647,  .647,  .647}{(0.10)} \\
          &       & \textcolor[rgb]{ .957,  .69,  .518}{\textbf{-9\%}} & \textcolor[rgb]{ .957,  .69,  .518}{\textbf{-8\%}} & \textcolor[rgb]{ .439,  .678,  .278}{\textbf{1\%}} & \textcolor[rgb]{ .957,  .69,  .518}{\textbf{-4\%}} \\
    \hdashline
    \multirow{3}[2]{*}{\large Llama3.2-3b-inst} & 47.21 & 43.76 & 43.5  & 33.54 & 39.77 \\
          & \textcolor[rgb]{ .647,  .647,  .647}{(0.08)} & \textcolor[rgb]{ .647,  .647,  .647}{(0.13)} & \textcolor[rgb]{ .647,  .647,  .647}{(0.10)} & \textcolor[rgb]{ .647,  .647,  .647}{(0.03)} & \textcolor[rgb]{ .647,  .647,  .647}{(0.13)} \\
          &       & \textcolor[rgb]{ .957,  .69,  .518}{\textbf{-7\%}} & \textcolor[rgb]{ .957,  .69,  .518}{\textbf{-8\%}} & \textcolor[rgb]{ .957,  .69,  .518}{\textbf{-29\%}} & \textcolor[rgb]{ .957,  .69,  .518}{\textbf{-16\%}} \\
    \bottomrule
    \end{tabular}%
    }
  \label{tab:Delimiter-sensitivity}%
\end{table*}%

\begin{table*}[h]
\centering
\caption{\small{Latency and Cost Analysis: PromptPrism increases total prompt length by 38.30\% over the baseline, which is more modest than competing methods: the OpenAI metaprompt approach \cite{openaimetaprompt},144.97\%, and the few-shot baseline,125.58\%. Critically, it achieves the highest performance score of 50.58, representing a 239.46\% improvement over baseline, delivering 6.25 pp of performance improvement per percentage point of length increase.}}
\resizebox{\textwidth}{!}{%
\begin{tabular}{l|r|r|r|r|r|r|}
\hline
                          & \multicolumn{1}{l}{Prompt Length (TOTAL)} & \multicolumn{1}{l|}{Prompt Length $\Delta$ (baseline)\%} & \multicolumn{1}{l|}{Prompt Length (INPUT)} & \multicolumn{1}{l|}{Prompt Length (OUTPUT)} & \multicolumn{1}{l}{Performance} & \multicolumn{1}{l|}{Performance $\Delta$ (baseline)\%} \\ \hline
Baseline                  & 1078.99                                   & \multicolumn{1}{l|}{}                             & 535.67                                    & 543.32                                     & 14.90                           & \multicolumn{1}{l|}{}                           \\
Baseline Fewshot 2        & 2433.97                                   & 125.58                                           & 2021.90                                   & 412.07                                     & 45.32                           & 204.16                                         \\
CoT                       & 718.54                                    & -33.41                                           & 688.68                                    & 29.85                                      & 41.29                           & 177.11                                         \\
OpenAI metaprompt \cite{openaimetaprompt} & 2643.19                                   & 144.97                                           & 2154.58                                   & 488.61                                     & 42.28                           & 183.76                                         \\
Ours                      & 1492.21                                   & 38.30                                            & 1429.68                                   & 62.53                                      & 50.58                           & 239.46 \\ \hline                                       
\end{tabular}
}
\label{tab:latency_cost}
\end{table*}%

\section{Related Work: Structured Prompts}\label{related_work_structured_prompts}
Recent research has made significant strides in understanding prompts as structured entities. Several approaches have emerged to analyze and categorize prompt components, from basic decomposition into instructions, examples, and queries \cite{fernando2023promptbreeder}, to more comprehensive analyses of real-world industrial prompt patterns \cite{mao2025prompts}. Some frameworks have proposed viewing prompts as programmable structures \cite{schnabel2024symbolic,wang2024langgpt}, drawing parallels with software development patterns and structured coding approaches. Others have explored structured prompts enhanced with graph-based representations to improve reasoning capabilities \cite{cheng2024structure}, while some work has focused on identifying recurring prompt patterns analogous to software design patterns \cite{white2023prompt}.

While these efforts contribute valuable insights into prompt structuring, they often lack comprehensive coverage of prompt characteristics. For example, frameworks such as \cite{fernando2023promptbreeder,wang2024langgpt} focus primarily on content-level analysis, overlooking the crucial aspects of stylistic formatting and syntactic variations. Our work addresses this gap by introducing a linguistically-inspired taxonomy that integrates multiple dimensions of prompt analysis. \textsc{PromptPrism} provides a comprehensive framework that encompasses stylistic formatting, syntactic patterns, and structural organization, enabling a more comprehensive understanding of how different prompt characteristics influence model behavior.

\section{Prompt Sensitivity Analysis}\label{prompt-sensitivity-code}
Fig \ref{fig:reorder-component} and Fig \ref{fig:delimiter-mod}, respectively shows the snippet of code, running controlled experiment on \textsc{PromptPrism} framework. The framework enables to conveniently run the semantic level intervention and syntactic modification experiment.

\section{Latency and Cost Analysis}\label{latency-cost-analysis}
Our proposed PromptPrism approach may introduce increased token count and computational costs. In Table \ref{tab:latency_cost} we present a comprehensive analysis comparing PromptPrism against other approaches, and the results show while there are tradeoffs between prompt length and performance gain, our approach show higher performance gain efficiency.



\begin{figure*}[h]
\begin{minted}
[frame=lines,
framesep=2mm,
%baselinestretch=1.2,
bgcolor=LightGray,
fontsize=\footnotesize,
%linenos
]
{python}
def reorder_component(self, component_name, position="first", remove_tags=False):
        """
        Reorder a specific component in the prompt.
        Args:
            component_name (str): Name of the component to reorder
            position (str): Where to place the component ('first', 'last','middle')

        Returns:
            str: Reordered prompt text
        """
        try:
            # Validate component name
            valid_components = ["request_query", "contextual_ref", "instruction", "output_const", "response", "other"]
            if component_name not in valid_components:
                raise ValueError(f"Invalid component name: {component_name}. Valid options are: {valid_components}")

            # Validate position
            validate_positions = ["first", "last", "middle"]
            if position not in validate_positions:
                raise ValueError(f"Invalid position: {position}. Valid options are: {validate_positions}")

            # find all tags that belong to the component category
            related_tags = [comp["index"] for comp in self.components if comp["tag"].startswith(f"{component_name}")]

            if not related_tags:
                raise ValueError(f"No components found for category: {component_name}")

            # remove related tags from current order
            other_tags = [comp["index"] for comp in self.components if not comp["tag"].startswith(f"{component_name}")]

            # Create new order based on position
            if position == "first":
                new_order = related_tags + other_tags
            elif position == "last":
                new_order = other_tags + related_tags
            else:  # middle
                mid_point = len(other_tags) // 2
                new_order = other_tags[:mid_point] + related_tags + other_tags[mid_point:]

            # update tag order
            self._tag_order = new_order
            # Use existing to_text()
            return self.to_text(remove_tags=remove_tags)

        except Exception as e:
            print(f"Error reordering component: {str(e)}")
            return self.to_text(remove_tags=remove_tags)
\end{minted}
\caption{Prompt sensitivity analysis: Semantic components re-ordering experiment implementation.}
\label{fig:reorder-component}
\end{figure*}%

\begin{figure*}[h]
\begin{minted}
[frame=lines,
framesep=2mm,
%baselinestretch=1.2,
bgcolor=LightGray,
fontsize=\footnotesize,
%linenos
]
{python}
def modify_delimiter(self, new_delimiter, position="all", remove_tags=False):
        """
        Modify delimiters between components based on specified position.
        Args:
            new_delimiter (str): The new delimiter to insert
            position (str): Where to modify delimiters ('all', 'first', 'last','middle')

        Returns:
            str: Updated prompt text with modified delimiters
        """
        if position not in ["all", "first", "last", "middle"]:
            raise ValueError(f"Invalid position: {position}. Valid options are: 'all', 'first', 'last', 'middle'")

        # Get number of components
        num_components = len(self.components)
        if num_components <= 1:
            return self.to_text()  # No delimiters to modify

        # Determine which indices to modify
        if position == "all":
            indices_to_modify = range(num_components - 1)
        elif position == "first":
            indices_to_modify = [0]
        elif position == "last":
            indices_to_modify = [num_components - 2]  # Second to last component
        elif position == "middle":
            if num_components >= 3:
                middle_idx = (num_components - 1) // 2
                indices_to_modify = [middle_idx]

        # Modify delimiters for specified indices
        for idx in indices_to_modify:
            self.components[idx]["delimiter_after"] = new_delimiter
            # Update metadata for the delimiter
            # self.components[idx]["metadata"]["delimiter_info"] = self._analyze_delimiter(new_delimiter)

        return self.to_text(remove_tags=remove_tags)
\end{minted}
\caption{Prompt Sensitivity Analysis: Syntactic Delimiter modification experiment implementation}
\label{fig:delimiter-mod}
\end{figure*}%
\clearpage%
\clearpage%

\section{Prompts Configuration}\label{prompt-app}

\begin{tcolorbox}[
    enhanced,
    colback=red!5!white,
    colframe=red!75!black,
    fonttitle=\bfseries,
    title={Prompt for taxonomy guided prompt generation},
    width=\textwidth,
    height=24cm,
]
\small{
\begin{Verbatim}
Your task is to augment the prompt based on the given prompt. 
Use these tags to augment the prompts:

1. <instruction> </instruction>: Prompt related to instructions to the LLM on what to do, guide 
how to think about or approach the problem. 
    1-1: <instruction:task> </instruction:task>: Task related instruction. 
        E.g., "This is a classification task."
    1-2: <instruction:guideline> </instruction:guideline>: Non-task specific instructions that 
            shape response behavior.  
        1-2-1: <instruction:guideline:role> </instruction:guideline:role>: Role assumption. 
            For example, "act as an experienced data scientist". 
        1-2-2: <instruction:guideline:scenario> </instruction:guideline:scenario>: 
            Senario description 
        1-2-3: <instruction:guideline:behavioral> </instruction:guideline:behavioral>: 
            Behavioral instructions and interaction style 
        1-2-4: <instruction:guideline:emotion> </instruction:guideline:emotion>:
            guideline for emotional tone
        1-2-5: <instruction:guideline:cot> </instruction:guideline:cot>: 
            Chain of Thought guideline. E.g., "Let's think step by step."
        1-2-6: <instruction:guideline:safety> </instruction:guideline:safety>:
            guideline related to safety. E.g., "avoid harmful content".
2. <contextual_ref> </contextual_ref>: Reference data of background information of the user query, 
    chat history, or retrieved sections in RAG. 
    2-1: <contextual_ref:fewshot> </contextual_ref:fewshot>: sample input-output pairs 
        for in-context learning.
    2-2: <contextual_ref:knowledge_base> </contextual_ref:knowledge_base>:
        reference information or facts. E.g., "Based on medical guidelines..."
    2-3: <contextual_ref:context_for_task> </contextual_ref:context_for_task>:
        relevant background information. 
3. <request_query> </request_query>: User request of query (e.g. Question) 
4. <output_const> </output_const>: Output constraints, response requirements to the LLM on 
    how to generate response 
    4-1: <output_const:label> </output_const:label>: defined set of possible output categories. 
        E.g., "choose from ['positive', 'negative']"
    4-2: <output_const:wordlimit> </output_const:wordlimit>: restrictions on response length. 
        E.g., "respond in 50 words or less".
    4-3: <output_const:format> </output_const:format>: output format or structure specification. 
        E.g., "format as JSON", "present in bullet points".
    4-4: <output_const:style_tone> </output_const:style_tone>: writing style or tone.
        E.g., "use academic language".
5. <other> </other>: Other purpose components
    5-1: <other:adversarial> </other:adversarial>: Adversarial components. E.g., "\&\&\&!!!!!!"
6. <response> </response>: Response component from LLM 
    6-1: <response:answer> </response:answer>: specific answer component.
7. <tools> </tools>: specifications for tool usage
    7-1: <tools:tool_name> </tools:tool_name>:identifier for specific tools
    7-2: <tools:tool_description> </tools:tool_description>: explanations of tool functionality. 
    7-3: <tools:parameters> </tools:parameters>: required inputs and configurations.

Be creative, and augment contents related to the tags. 
For example, <output\_const> Give only the answer </output\_const>.
Give **ONLY** the augmented version of the prompt. Do **NOT** include any preambles.
                              
Below is the prompt:

### Prompt ### 
<instruction> {definition} </instruction>}

### Examples ###

### Negative Examples ### 
<contextual_ref> {negative_examples} </contextual_ref>

### Positive Examples ###
<contextual_ref> {positive_examples} </contextual_ref>
\end{Verbatim}
}
\end{tcolorbox}
\clearpage

\newpage
\begin{PromptBox}
\small{
\begin{Verbatim}
You task is to annotate a prompt based on some prompt taxonomy defined below. 

Given a prompt, please decompose the prompt into several components by adding the below tags in 
prompt taxonomy section to the original input prompt. 
The available prompt component tags are defined below. Not all prompt component tags should be 
used for each prompt. 

### prompt taxonomy ###
{prompt_taxonomy}

### additional tips ###
1. You have to analyze system prompts and user prompts separately using the above prompt taxonomy.

2. If you can find the related tags in a subcategory component, use the tag in the subcategory. 
If not, use the parent node tag. For example, if a prompt segment belongs to few-shot examples, 
then use "<contextual_ref:fewshot> </contextual_ref:fewshot>", then insert this tag. 
If it belongs to other contextual information that are not directly available in the options, 
then use "<contextual_ref> </contextual_ref>" instead.

3. Please only added tags to the original input prompt. Please do **NOT** include any preambles 
or additional explanations. 

4. If there already exists some tags in the original input prompt such as <tools></tools>, 
think about if the tags align with our current taxonomy defined above. If the tag name is the 
same, keep the original tags without adding additional tags. If the tag name is different, 
insert an additional tag outside of the original tag. 

5. It is possible to have the same tag to be used more than once in a single input prompt. 
However, for consecutive prompt components sharing the same tags, merge them and use only one tag 
instead of multiple same tags. 

### few-shot examples ###
Input prompt: {example_input_prompt}
Output: {example_output}

### Prompt ###
This is the input prompt you want to annotate: {input_prompt_to_be_annotated}
\end{Verbatim}
}
\end{PromptBox}
\clearpage
\newpage

\begin{tcolorbox}[
    breakable,
    enhanced,
    colback=red!5!white,
    colframe=red!75!black,
    fonttitle=\bfseries,
    title={Prompt for task annotation},
    width=\textwidth,
    listing only, 
    listing engine=listings,
    listing options={
    basicstyle=\ttfamily\small,
    breaklines=true,
    breakatwhitespace=false,
    postbreak=\mbox{\textcolor{red}{$\hookrightarrow$}\space},
    columns=fullflexible,
    keepspaces=true,
    showstringspaces=false
  }
]

\small{
\begin{Verbatim}
You are an LLM prompt task type classifier. Your task is to classify a prompt to certain task type. 
Please classify the prompt into one of the below task types:
### Prompt Task Type Options ###
* Classification:Sentiment Analysis
* Classification:Topic Classification
* Classification:Toxicity Detection
* Classification:Multi-label Classification
* Classification:Others
* Open Book QA:Reading Comprehension
* Open Book QA:Document-based QA
* Open Book QA:Multi-document QA
* Open Book QA:Context-specific Questions
* Open Book QA:Others
* Closed Book QA:Factual QA
* Closed Book QA:Common Knowledge Questions
* Closed Book QA:Others
* Coding:Algorithm Implementation
* Coding:Debugging
* Coding:Code Generation
* Coding:Code Explanation
* Coding:Function Implementation
* Coding:Others
* Reasoning:Logical Reasoning
* Reasoning:Mathematical Problem Solving
* Reasoning:Causal Reasoning
* Reasoning:Others
* Summarization:Text Summarization
* Summarization:Multi-document Summarization
* Summarization:Others
* Function calling:Data Retrieval (e.g., get_user_info, fetch_record_by_id)
* Function calling:System Operations (e.g., check_status, validate_token)
* Function calling:API Integration (e.g., external_api_call, service_request)
* Function calling:Data Transformation (e.g., format_data, convert_units)
* Function calling:Multi-function Chain (e.g.,complex operations of multiple function calls)
* Function calling:Others
* Others
### Additional Tips ###
1. If a prompt belongs to logical reasoning of the reasoning task type, then your output should be 
  "Reasoning:Logical Reasoning". 
2. Please use the uppercase letter and Lowercase letter exactly as above. Please do not include 
   any information in the parenthesis in the task type name. 
3. If you cannot classify a prompt into a task type, please feel free to use "Others". For 
   example, if a task belongs to summarization, but not text summarization or multi-doc 
   summarization, then the output should be "Summarization:Others". 
4. Please only output task type of the prompt. Please do **NOT** include any preambles or 
   additional explanations. 
5. Please do not output more than one task type for a given prompt. Each prompt should only be 
   linked to one task type. 
6. Typically, if a prompt contains one or more tools, you may probably classify it into one of 
   the function calling task types.
7. Please do **NOT** extract actual content from the prompt as task type. Please only use the 
   above options as the task type. 

### Few-shot Examples ###
Input Prompt: "{example task instruction}"
Output: "{task type}"
### Prompt ###
This is the input prompt you want to annotate: {input_prompt_to_be_annotated}
\end{Verbatim}
}
\end{tcolorbox}

\label{prompt: annotation}
\clearpage

\newpage

\newtcolorbox{PromptBox-blue}{
  breakable,
  enhanced,
  colback=blue!5!white,
  colframe=blue!75!black,
  fonttitle=\bfseries,
  title={Prompt for taxonomy annotation},
  width=\textwidth,
  listing only,
  listing engine=listings,
  listing options={
    basicstyle=\ttfamily\small,
    breaklines=true,
    breakatwhitespace=false,
    postbreak=\mbox{\textcolor{red}{$\hookrightarrow$}\space},
    columns=fullflexible,
    keepspaces=true,
    showstringspaces=false
  }
}

\begin{tcolorbox}[
    breakable,
    enhanced,
    colback=blue!5!white,
    colframe=blue!75!black,
    fonttitle=\bfseries,
    title={Example of taxonomy guided prompt},
    width=\textwidth,
    listing only, 
    listing engine=listings,
    listing options={
    basicstyle=\ttfamily\small,
    breaklines=true,
    breakatwhitespace=false,
    postbreak=\mbox{\textcolor{red}{$\hookrightarrow$}\space},
    columns=fullflexible,
    keepspaces=true,
    showstringspaces=false
  }
]
\small{
\begin{Verbatim}
<instruction:task>Classify the given tweet into one of three categories: Hate Speech, Offensive, 
or Normal. Analyze the content for threatening, abusive, or discriminatory language towards 
specific communities.</instruction:task>

<instruction:guideline:behavioral>Approach each tweet objectively and systematically. 
Consider the language used, the target of the message, and the overall tone.
</instruction:guideline:behavioral>

<instruction:guideline:safety>Be aware that you may encounter disturbing or offensive content. 
Maintain a professional demeanor and focus on the classification task.
</instruction:guideline:safety>

<contextual_ref:fewshot>
{omitted due to limited space}
</contextual_ref:fewshot>

<output_const:format>Respond with the classification (Hate Speech, Offensive, or Normal) 
followed by a brief explanation.</output_const:format>

<output_const:style_tone>Maintain a neutral and analytical tone in your response.
</output_const:style_tone>

<other:adversarial>Be cautious of tweets that may seem offensive at first glance but do not 
actually target any specific community.</other:adversarial>
\end{Verbatim}
}
\end{tcolorbox}
\label{example-taxonomy-guided}

\begin{tcolorbox}[
    breakable,
    enhanced,
    colback=blue!5!white,
    colframe=blue!75!black,
    fonttitle=\bfseries,
    title={Example of taxonomy annotated prompt: apigen-80k },
    width=\textwidth,
    listing only, 
    listing engine=listings,
    listing options={
    basicstyle=\ttfamily\small,
    breaklines=true,
    breakatwhitespace=false,
    postbreak=\mbox{\textcolor{red}{$\hookrightarrow$}\space},
    columns=fullflexible,
    keepspaces=true,
    showstringspaces=false
  }
]
\small{
\begin{Verbatim}
<instruction:guideline:role>You are an expert in composing functions.
</instruction:guideline:role>

<instruction:guideline>
You are given a question and a set of possible functions. 
Based on the question, you will need to make one or more function/tool calls to achieve the
purpose. If none of the functions can be used, point it out and refuse to answer. 
If the given question lacks the parameters required by the function, also point it out.
</instruction:guideline>

<tools:tool_description>You have access to the following tools:</tools:tool_description>
<tools>[{"type":"function","function":{"name":"similarity_score","description":
"Calculates the similarity score between two lists of integers.",
"parameters":{"type":"object",
"properties":{"list1":{"type":"array","items":{"type":"integer"},
"description":"The first list of integers."},
"list2":{"type":"array","items":{"type":"integer"},
"description":"The second list of integers."}},
"required":["list1","list2"]}}}]</tools>

<output_const:format>The output MUST strictly adhere to the following format, and NO other text 
MUST be included.
The example format is as follows. Please make sure the parameter type is correct. 
If no function call is needed, please make the tool calls an empty list '[]'.
<tool_call>[
{"name": "func_name1", "arguments": {"argument1": "value1", "argument2": "value2"}},
... (more tool calls as required)
]</tool_call></output_const:format>

<request_query>
Calculate the similarity score between the lists 
[1, 2, 3, 4, 5] and [3, 4, 5, 6, 7].
</request_query>
\end{Verbatim}
}
\end{tcolorbox}
\label{example-taxonomy-guided}
\clearpage
\newpage